# Change in Abstract Argumentation Frameworks: Adding an Argument

**Claudette Cayrol**                                                    CCAYROL@IRIT.FR
**Florence Dupin de Saint-Cyr**                                        BANNAY@IRIT.FR
**Marie-Christine Lagasquie-Schiex**                                   LAGASQ@IRIT.FR
*IRIT, Université Paul Sabatier,*
*118 route de Narbonne, 31062 Toulouse, France*

## Abstract

In this paper, we address the problem of change in an abstract argumentation system. We focus on a particular change: the addition of a new argument which interacts with previous arguments. We study the impact of such an addition on the outcome of the argumentation system, more particularly on the set of its extensions. Several properties for this change operation are defined by comparing the new set of extensions to the initial one, these properties are called "structural" when the comparisons are based on set-cardinality or set-inclusion relations. Several other properties are proposed where comparisons are based on the status of some particular arguments: the accepted arguments; these properties refer to the "evolution of this status" during the change, e.g., Monotony and Priority to Recency. All these properties may be more or less desirable according to specific applications. They are studied under two particular semantics: the grounded and preferred semantics.

## 1. Introduction

Argumentation has become an influential approach to handle Artificial Intelligence problems including defeasible reasoning (see e.g., Pollock, 1992; Dung, 1995; Bondarenko, Dung, Kowalski, & Toni, 1997; Chesñevar, Maguitman, & Loui, 2000; Prakken & Vreeswijk, 2002; Amgoud & Cayrol, 2002; Nute, 2003), and modeling agents interactions (see e.g., Amgoud, Maudet, & Parsons, 2000; Kakas & Moraïtis, 2003). Argumentation is basically concerned with the exchange of interacting arguments. This set of arguments may come either from a dialogue between several agents but also from the available (and possibly contradictory) pieces of information at the disposal of one unique agent. Usually, the interaction between arguments takes the form of a conflict, called attack. For example, a logical argument can be a pair ⟨set of assumptions, conclusion⟩, where the set of assumptions entails the conclusion according to some logical inference schema. A conflict occurs, for instance, when the conclusion of an argument contradicts an assumption of another argument.

The main issue for any argumentation system is the selection of acceptable sets of arguments, called "extensions", based on the way arguments interact (intuitively, an acceptable set of arguments must be in some sense coherent and strong enough, e.g., able to defend itself against all attacking arguments). So, the outcome of an argumentation system is often defined by the set of its extensions but, depending on the applications, it may be also defined as the set of arguments that belongs to every extension. It is convenient to explore the concept of extension through argumentation frameworks, and especially Dung's (1995) framework, which abstracts from the arguments nature, and represents interaction under the form of a binary relation "attack" on a set of arguments.

Recent works have considered the dynamics of such abstract argumentation frameworks (Cayrol, Dupin de Saint-Cyr, & Lagasquie-Schiex, 2008; Rotstein, Moguillansky, García, & Simari, 2008b; Boella, Kaci, & van der Torre, 2009a, 2009b). The problem is to study how the outcome changes when the set of arguments and/or the set of attacks between them are changed. In this paper, we focus on the case when a new argument and its interactions are added to an argumentation system. We study the impact of such an addition on the set of initial extensions. This leads us to identify some properties of the change operation with respect to the modification it induces on the outcome. This study has two main applications, the first one concerns computational issues, while the second one concerns the definition of dialogue strategies. On one hand, the interest for computational processing is that knowledge about the properties of the change may help to deduce what are the modifications in the extensions. For instance, it is useful to know conditions under which change will not modify the previous extensions. On the other hand, knowing the impact of adding an argument may help choosing the good one in order to achieve a given goal. For instance, in a multi-agent setting, *i.e.*, when several agents may present several arguments, the results presented in this paper will help one agent to determine which arguments she should present in order that the outcome of the dialogue satisfies desired properties. For example, if she wants to widen the debate, the argument that must be added should induce a change producing larger extensions (*i.e.* that contain more arguments, see Section 3 and Section 5).

The paper is organized as follows. Section 2 recalls the basic concepts in argumentation. Section 3 settles a definition of change in argumentation. Many features can be taken into account in order to characterize a change operation. We first propose a class of properties based on the impact of the change on the structure of the resulting set of extensions (see Section 3.2). In a second step, we define several other properties regarding the arguments themselves, particularly those which are accepted after change (see Section 3.3). These properties are defined regardless of the semantics.

Then, we focus on a particular change: the addition of a new argument which may interact with previously introduced arguments. Section 4 is dedicated to the study of the properties of this addition in the case of two particular semantics, the grounded and the preferred semantics. We give conditions under which a given property is satisfied. Section 5 discusses the related approaches in the literature. All the proofs (and two important lemmas) are given in Appendix A. Some additional examples are presented in Appendix B for illustrating the other change operations.

Note that this paper generalizes a previous work (Cayrol et al., 2008), where argument addition, called "revision", was restricted to one argument having only one interaction with the existing argumentation system. Here, the added argument may interact with *any number of previous arguments*. Moreover, a broader analysis of this generalized addition is provided by considering *new properties* such as, e.g., Monotony, and by establishing *new connections* between the different properties.

## 2. Basic Concepts in Argumentation Frameworks

The present work lies in the frame of the general theory of abstract argumentation frameworks proposed by Dung (1995). Such an abstract framework assumes that a set of arguments is given, as well as the different conflicts between them, and focuses on the definition of the status of arguments.

**Definition 1 (Argumentation framework)** *An argumentation framework $\langle \mathbf{A}, \mathbf{R} \rangle$ is a pair, where $\mathbf{A}$ is a non-empty set and $\mathbf{R}$ is a binary relation on $\mathbf{A}$, called attack relation. Let $A, B \in \mathbf{A}$, $(A, B) \in \mathbf{R}$ or equivalently $A\mathbf{R}B$ means that $A$ attacks $B$, or $B$ is attacked by $A$.*

In the following, $\langle \mathbf{A}, \mathbf{R} \rangle$ is an argumentation framework, and we assume that the set of arguments $\mathbf{A}$ is finite. First, it is easy to extend the concept of attack to sets of arguments.

**Definition 2 (Attack from and to a set)** *Let $A \in \mathbf{A}$ and $\mathcal{S} \subseteq \mathbf{A}$.[1]*

- *$\mathcal{S}$ attacks $A$ iff $\exists X \in \mathcal{S}$ such that $X\mathbf{R}A$.*

- *$A$ attacks $\mathcal{S}$ iff $\exists X \in \mathcal{S}$ such that $A\mathbf{R}X$.*

The main issue of any argumentation system is the selection of acceptable sets of arguments. Intuitively, an acceptable set of arguments must be in some sense coherent and strong enough (e.g., able to defend itself against every attacking argument). An argumentation semantics defines the properties required for a set of arguments to be acceptable (this is a collective acceptability). The selected sets of arguments under a given semantics are called extensions of that semantics. The set of extensions characterizes the outcome of an argumentation system. We recall the basic concepts used for defining usual semantics:

**Definition 3 (Conflict-free, defense)** *Let $A \in \mathbf{A}$ and $\mathcal{S} \subseteq \mathbf{A}$.*

- *$\mathcal{S}$ is conflict-free iff $\nexists A, B \in \mathcal{S}$ such that $A\mathbf{R}B$.*

- *$\mathcal{S}$ defends $A$ iff $\mathcal{S}$ attacks each argument which attacks $A$. The set of arguments which $\mathcal{S}$ defends will be denoted by $\mathcal{F}(\mathcal{S})$. $\mathcal{F}$ is called the characteristic function of $\langle \mathbf{A}, \mathbf{R} \rangle$.*

The literature proposes an increasing variety of semantics, refining Dung's traditional ones (Baroni, Giacomin, & Guida, 2005; Caminada, 2006; Dung, Mancarella, & Toni, 2006; Coste-Marquis, Devred, & Marquis, 2005). In this paper, only the most well-known traditional semantics are considered: the grounded, preferred and stable semantics.

**Definition 4 (Acceptability semantics)** *Let $\mathcal{E} \subseteq \mathbf{A}$.*

- *$\mathcal{E}$ is admissible iff $\mathcal{E}$ is conflict-free and defends all its elements (i.e. $\mathcal{E} \subseteq \mathcal{F}(\mathcal{E})$).*

- *$\mathcal{E}$ is a preferred extension iff $\mathcal{E}$ is a maximal (w.r.t. set-inclusion) admissible set.*

---

1. In this paper, we use $\subset$ to denote strict inclusion and $\subseteq$ to denote classical inclusion.

- $\mathcal{E}$ is the grounded extension *iff $\mathcal{E}$ is the least fixed point (w.r.t. set-inclusion) of the characteristic function $\mathcal{F}$.*

- $\mathcal{E}$ is a stable extension *iff $\mathcal{E}$ is conflict-free and attacks each argument which does not belong to $\mathcal{E}$.*

An argumentation framework can be represented as a directed graph, called attack graph, where nodes are the arguments and edges represent the attack relation. Throughout the paper, examples are using this graph representation.

**Example 1**
$\mathbf{A} = \{A, B, C, D, F\}$ *and* $\mathbf{R} = \{(A, B), (B, A), (B, C), (C, D), (D, F), (F, C)\}$.

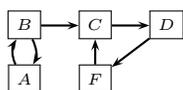

*The admissible sets are* $\{\}$, $\{A\}$, $\{B\}$ *and* $\{B, D\}$.
*The preferred extensions are* $\{A\}$ *and* $\{B, D\}$.
*The grounded extension is* $\{\}$.
$\{B, D\}$ *is the unique stable extension.*

Using the graph-based representation of an argumentation framework, we extend the definition of individual attack as follows:

**Definition 5 (indirect attack and defense)** *Let $\mathcal{G}$ denote the attack graph associated with $\langle \mathbf{A}, \mathbf{R} \rangle$. Let $A$, $B \in \mathbf{A}$.*

- *A indirectly attacks B iff there is an odd-length path from A to B in the attack graph $\mathcal{G}$.*

- *A indirectly defends B iff there is an even-length path (with non-zero length) from A to B in the attack graph $\mathcal{G}$.*

Note that the case when $A$ attacks $B$ is considered as a particular case of indirect attack. Dung (1995) has proved the following results.

**Proposition 1** *Let $\langle \mathbf{A}, \mathbf{R} \rangle$ be an argumentation framework.*

1. *There is at least one preferred extension, always a unique grounded extension, while there may be zero, one or many stable extensions.*

2. *Each admissible set is included in a preferred extension.*

3. *Each stable extension is a preferred extension, the converse is false.*

4. *The grounded extension is included in each preferred extension.*

5. *Each argument which is not attacked belongs to the grounded extension (hence to each preferred and to each stable extension).*

6. *If $\mathbf{R}$ is finite, the grounded extension can be computed by iteratively applying the function $\mathcal{F}$ from the empty set.*

The presence of cycles in the attack graph has often raised some problems, namely for the stable semantics, for which it may happen that no extension exists. Note that some authors only consider attack graphs without odd-length cycles, arguing that an odd-length cycle carries counterintuitive information. The following results give properties of the preferred, grounded and stable extensions depending on the existence of cycles in the attack graph.

**Proposition 2** *(Dunne & Bench-Capon, 2001, 2002) Let $\mathcal{G}$ denote the attack graph associated with $\langle \mathbf{A}, \mathbf{R} \rangle$.*

1. *If $\mathcal{G}$ contains no cycle, $\langle \mathbf{A}, \mathbf{R} \rangle$ has a unique preferred extension, which is also the grounded extension and the unique stable extension.*

2. *If $\{\}$ is the unique preferred extension of $\langle \mathbf{A}, \mathbf{R} \rangle$, $\mathcal{G}$ contains an odd-length cycle.*

3. *If $\langle \mathbf{A}, \mathbf{R} \rangle$ has no stable extension, $\mathcal{G}$ contains an odd-length cycle.*

4. *If $\mathcal{G}$ contains no odd-length cycle, preferred and stable extensions coincide.*

5. *If $\mathcal{G}$ contains no even-length cycle, $\langle \mathbf{A}, \mathbf{R} \rangle$ has a unique preferred extension.*

Now that acceptable sets of arguments have been defined, it is possible to define a status for an individual argument.

**Definition 6 (Argument status)** *Let $\langle \mathbf{A}, \mathbf{R} \rangle$ be an argumentation framework and $A \in \mathbf{A}$. Given a semantics $s$:*

- *$A$ is* skeptically accepted *under $s$ iff $A$ belongs to each extension of $\langle \mathbf{A}, \mathbf{R} \rangle$ under $s$.*

- *$A$ is* credulously accepted *under $s$ iff $A$ belongs to at least one extension of $\langle \mathbf{A}, \mathbf{R} \rangle$ under $s$.*

- *$A$ is* rejected *under $s$ iff $A$ does not belong to any extension of $\langle \mathbf{A}, \mathbf{R} \rangle$ under $s$.*

Obviously, credulous and skeptical acceptance coincide under the grounded semantics.

## 3. Change in Argumentation

We introduce a formal definition of change in argumentation which enables to distinguish between four types of change. Then we define properties for change in argumentation. First, we consider the impact of a change operation on the structure of the set of extensions, and we study how this structure is modified. This point of view leads to the definition of structural properties. Then, we consider the impact of a change operation on the set of arguments which are accepted. Finally, the connections between both classes of properties are studied.

Note that for most of the properties that we introduce, the definition is general in the sense that it can be applied to any type of change. In Section 4 (where we give conditions for satisfying these properties), we will focus on the particular case of the addition of an argument and its interactions.

### 3.1 Definition

In this section, we give a definition of change in argumentation. The change may concern the set of arguments and/or the set of attacks between them. So, at least four cases can be encountered:

**Definition 7 (Change operations)** *Let $\langle \mathbf{A}, \mathbf{R} \rangle$ be an argumentation framework.*

- adding only one interaction $i_0$ *between two existing arguments of* $\mathbf{A}$ *($i_0 = (X, Y)$ with $X \in \mathbf{A}$ and $Y \in \mathbf{A}$) is a change operation defined by:*

$$\langle \mathbf{A}, \mathbf{R} \rangle \oplus_i i_0 = \langle \mathbf{A}, \mathbf{R} \cup \{i_0\} \rangle$$

- removing only one existing interaction $i_0$ *of* $\langle \mathbf{A}, \mathbf{R} \rangle$ *($i_0 \in \mathbf{R}$) is a change operation defined by:*

$$\langle \mathbf{A}, \mathbf{R} \rangle \ominus_i i_0 = \langle \mathbf{A}, \mathbf{R} \setminus \{i_0\} \rangle$$

- adding only one argument $Z \notin \mathbf{A}$ *and a set of interactions concerning* $Z$ *denoted by* $\mathcal{I}_z$ *is a change operation defined by:*

$$\langle \mathbf{A}, \mathbf{R} \rangle \oplus_i^a (Z, \mathcal{I}_z) = \langle \mathbf{A} \cup \{Z\}, \mathbf{R} \cup \mathcal{I}_z \rangle$$

  *Here, $\mathcal{I}_z$ is supposed to be a non-empty set of pairs of arguments (either of the form $(X, Z)$ or $(Z, X)$ with $X \in \mathbf{A})^2$*

- removing only one argument $Z \in \mathbf{A}$ *which interacts with other arguments is a change operation defined by:*

$$\langle \mathbf{A}, \mathbf{R} \rangle \ominus_i^a Z = \langle \mathbf{A} \setminus \{Z\}, \mathbf{R} \setminus \mathcal{I}_z \rangle$$

  *Here, $\mathcal{I}_z$ denotes the set of all the interactions concerning $Z$, that is the set $\{(Z, X) \mid (Z, X) \in \mathbf{R}\} \cup \{(X, Z) \mid (X, Z) \in \mathbf{R}\}^3$*

Note that the case of adding a new argument (resp. removing an existing argument) which does not interact with any other argument is trivial: it has only to be added to (resp. removed from) each extension. Indeed, change is more interesting when the concerned argument interacts with previous ones.

In a very recent work about dynamics of argumentation (Boella et al., 2009a, 2009b), the four types of change defined above have been introduced under different names, respectively attack refinement, attack abstraction, argument refinement and argument abstraction. However, only the operations of attack refinement, attack abstraction and argument abstraction have been studied and in a more restricted context (see Section 5 for a discussion).

In the following, we identify an argumentation framework $\langle \mathbf{A}, \mathbf{R} \rangle$ with its associated attack graph $\mathcal{G}$. We write $X \in \mathcal{G}$ instead of "$X$ is an argument represented by a node of $\mathcal{G}$". The set of extensions of $\langle \mathbf{A}, \mathbf{R} \rangle$ is denoted by $\mathbf{E}$ (with $\mathcal{E}_1, \ldots, \mathcal{E}_n$ denoting the extensions).

---

2. Note that, by this definition, it is impossible to have $(Z, Z)$ in $\mathcal{I}_z$.
3. Note that if $Z$ is removed, the set of interactions concerning $Z$ must be also removed.

A change operation produces a new framework $\langle \mathbf{A}', \mathbf{R}' \rangle$ represented by a graph $\mathcal{G}'$, with a new set of extensions $\mathbf{E}'$ (with $\mathcal{E}'_1, \ldots, \mathcal{E}'_p$ denoting the extensions).

As explained above, changing an argumentation framework may modify the set of extensions. Given a semantics, the modifications are more or less important. It depends on the kinds of interactions that are added or removed and more precisely on the status of the arguments involved in these interactions.

The impact of a change can be studied from two points of view:

- the first one concerns the *structure of the set of extensions* and it can address either the comparison of the number of extensions before and after the change, or, if this number remains unchanged, the comparison of the contents of the extensions before and after the change;

- the second point of view concerns the *status of some particular arguments*.

So, in the next sections, we propose two classes of general properties for a change operation, one for each point of view. The proposed properties characterize the relation between a particular framework and the resulting framework after a change.

## 3.2 Structural Properties

Structural properties, presented in this section, are based on the impact of the change on the structure of the set of extensions. Note that for each property, the definition is general in the sense that the type of change operation is not specified: it can consist in adding one interaction, removing one interaction, adding an argument and a set of interactions concerning this argument, or removing one argument. However, for sake of clarity, each property will be illustrated in this section with examples for the change operation $\oplus_i^a$; the reader will find some examples for the other change operations in Appendix B.

Let $\langle \mathbf{A}, \mathbf{R} \rangle$ be an argumentation framework and $\mathbf{E}$ be the set of extensions of $\langle \mathbf{A}, \mathbf{R} \rangle$ (under a given semantics $s$). Various situations may be encountered in the general case. $\mathbf{E}$ may be empty (implying that $s$ is the stable semantics), may be reduced to a singleton $\{\mathcal{E}_1\}$ (where $\mathcal{E}_1$ may be empty), or may contain more than one extension $\{\mathcal{E}_1, \ldots, \mathcal{E}_n\}$. The situation with only one non-empty extension is convenient for the determination of the status of an argument. By contrast, when several extensions exist, different choices are available. Table 1 summarizes the various definitions presented below.

We first consider the **decisive** property for a change operation, meaning that $\mathcal{G}'$ has a unique non-empty extension, while it was not the case for $\mathcal{G}$.

**Definition 8 (Decisive change)** *The change from $\mathcal{G}$ to $\mathcal{G}'$ is decisive iff $\mathbf{E} = \varnothing$, or $\mathbf{E} = \{\{\}\}$, or $\mathbf{E} = \{\mathcal{E}_1, \ldots, \mathcal{E}_n\}$, $n \geq 2$, and $\mathbf{E}' = \{\mathcal{E}'\}$, $\mathcal{E}' \neq \{\}$.*

## Example 2

1. *Under the stable (resp. grounded or preferred) semantics, the change $\oplus_i^a$ with $Z$ and $\mathcal{I}_z = \{(Z, A)\}$ is decisive since:*

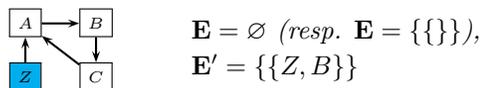

   $\mathbf{E} = \varnothing$ *(resp.* $\mathbf{E} = \{\{\}\}$),
   $\mathbf{E}' = \{\{Z, B\}\}$

| Property for a change operation | Characterization of the property |
|---|---|
| the change is **decisive** | $\mathbf{E} = \varnothing$ or $\mathbf{E} = \{\{\}\}$ or $|\mathbf{E}| > 2$ and $|\mathbf{E}'| = 1$ and $\mathbf{E}' \neq \{\{\}\}$ |
| the change is **restrictive** | $|\mathbf{E}| > |\mathbf{E}'| > 2$ |
| the change is **questioning** | $|\mathbf{E}| < |\mathbf{E}'|$ |
| the change is **destructive** | $\mathbf{E} \neq \varnothing$ and $\mathbf{E} \neq \{\{\}\}$ $\mathbf{E}' = \varnothing$ or $\mathbf{E}' = \{\{\}\}$ |
| the change is **expansive** | $|\mathbf{E}| = |\mathbf{E}'|$ and $\forall \mathcal{E}'_j \in \mathbf{E}', \exists \mathcal{E}_i \in \mathbf{E}, \mathcal{E}_i \subset \mathcal{E}'_j$ |
| the change is **conservative** | $\mathbf{E} = \mathbf{E}'$ |
| the change is **altering** | $|\mathbf{E}| = |\mathbf{E}'|$ and $\exists \mathcal{E}_i \in E$ s.t. $\forall \mathcal{E}'_j \in \mathbf{E}', \mathcal{E}_i \nsubseteq \mathcal{E}'_j$ |

Table 1: Structural properties for a change operation

2. *Under the grounded semantics, the change $\oplus_i^a$ with $Z$ and $\mathcal{I}_z = \{(Z, A)\}$ is decisive since:*

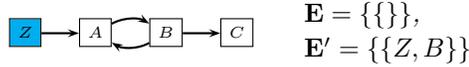
   $\mathbf{E} = \{\{\}\},$
   $\mathbf{E}' = \{\{Z, B\}\}$

3. *Under the preferred semantics, the change $\oplus_i^a$ with $Z$ and $\mathcal{I}_z = \{(Z, A)\}$ is decisive since:*

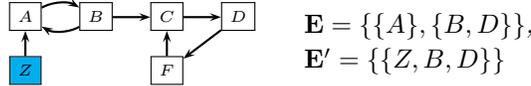
   $\mathbf{E} = \{\{A\}, \{B, D\}\},$
   $\mathbf{E}' = \{\{Z, B, D\}\}$

4. *Under the preferred semantics, the change $\oplus_i^a$ with $Z$ and $\mathcal{I}_z = \{(Z, A),\ (B, Z)\}$ is decisive since:*

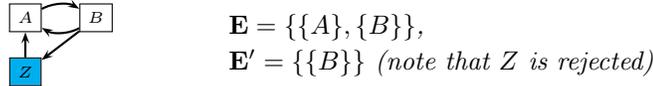
   $\mathbf{E} = \{\{A\}, \{B\}\},$
   $\mathbf{E}' = \{\{B\}\}$ *(note that $Z$ is rejected)*

A weaker requirement is the decrease of the number of choices. A change such that $\mathcal{G}'$ has strictly less extensions than $\mathcal{G}$, but still has at least two, is called **restrictive**[4]. Note that the restrictive property does not make sense under the grounded semantics, since there is always a unique grounded extension.

**Definition 9 (Restrictive change)** *The change from $\mathcal{G}$ to $\mathcal{G}'$ is restrictive iff $\mathbf{E} = \{\mathcal{E}_1,$ $\ldots, \mathcal{E}_n\}$, $n \geq 2$, and $\mathbf{E}' = \{\mathcal{E}'_1, \ldots, \mathcal{E}'_p\}$, with $n > p \geq 2$.*

**Example 3**

1. *Under the preferred (or stable) semantics, the change $\oplus_i^a$ with $Z$ and $\mathcal{I}_z = \{(Z, A)\}$ is restrictive since:*

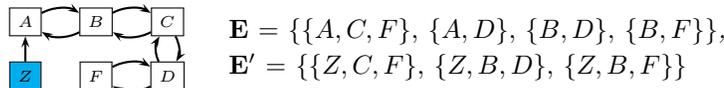
   $\mathbf{E} = \{\{A, C, F\}, \{A, D\}, \{B, D\}, \{B, F\}\},$
   $\mathbf{E}' = \{\{Z, C, F\}, \{Z, B, D\}, \{Z, B, F\}\}$

---

4. In the work of Cayrol et al. (2008), this kind of change was called "selective".

2. *Under the preferred semantics, the change $\oplus_i^a$ with $Z$ and $\mathcal{I}_z = \{(Z, A), (B, Z)\}$ is restrictive since:*

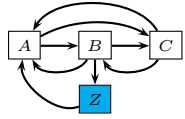

$\mathbf{E} = \{\{A\}, \{B\}, \{C\}\}$,
$\mathbf{E}' = \{\{B\}, \{C, Z\}\}$ *(note that $Z$ is not skeptically accepted)*

An opposite point of view enables to consider changes which raise ambiguity, by increasing the number of extensions. This is the case for instance when $\mathcal{G}$ has at least one non-empty extension and $\mathcal{G}'$ has strictly more extensions than $\mathcal{G}$. A slightly different situation occurs when $\mathcal{G}$ has no extension or an empty one, while $\mathcal{G}'$ has more than one extension. In that case, change brings some information, but is not decisive. Such changes are called **questioning**. As for the restrictive property, the questioning property does not make sense under the grounded semantics.

**Definition 10 (Questioning change)** *The change from $\mathcal{G}$ to $\mathcal{G}'$ is questioning iff $\mathbf{E}' = \{\mathcal{E}'_1, \ldots, \mathcal{E}'_p\}$, with $p \geq 2$, and either $\mathbf{E} = \varnothing$, or $\mathbf{E} = \{\mathcal{E}_1, \ldots, \mathcal{E}_n\}$ and $p > n \geq 1$.*

**Example 4**

1. *Under the preferred (or stable) semantics, the change $\oplus_i^a$ with $Z$ and $\mathcal{I}_z = \{(Z, A)\}$ is questioning since:*

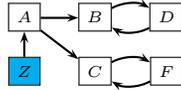

$\mathbf{E} = \{\{A, D, F\}\}$,
$\mathbf{E}' = \{\{Z, B, C\}, \{Z, B, F\}, \{Z, D, C\}, \{Z, D, F\}\}$

2. *Under the stable semantics, the change $\oplus_i^a$ with $Z$ and $\mathcal{I}_z = \{(Z, A)\}$ is questioning since:*

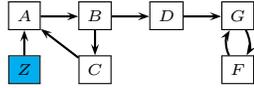

$\mathbf{E} = \varnothing$,
$\mathbf{E}' = \{\{Z, B, F\}, \{Z, B, G\}\}$

3. *Under the preferred semantics, the change $\oplus_i^a$ with $Z$ and $\mathcal{I}_z = \{(Z, A), (A, Z), (Z, B), (B, Z)\}$ is questioning since:*

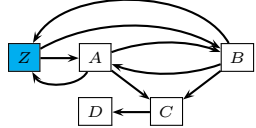

$\mathbf{E} = \{\{A, D\}, \{B, D\}\}$,
$\mathbf{E}' = \{\{A, D\}, \{B, D\}, \{Z\}\}$ *(note that $Z$ is not skeptically accepted)*

Pursuing along the previous line, we consider changes leading to a kind of decisional dead-end. This is the case when $\mathcal{G}$ has at least one non-empty extension and $\mathcal{G}'$ has no extension, or an empty one[5]. Such a change is called **destructive**.

**Definition 11 (Destructive change)** *The change from $\mathcal{G}$ to $\mathcal{G}'$ is destructive iff $\mathbf{E} = \{\mathcal{E}_1, \ldots, \mathcal{E}_n\}$, $n \geq 1$, $\mathcal{E}_i \neq \{\}$ and $\mathbf{E}' = \varnothing$ or $\mathbf{E}' = \{\{\}\}$.*

**Example 5**

---

5. These are two different cases but they have the same impact: there is no possible decision because no argument is accepted.

1. *Under the stable semantics, the change $\oplus_i^a$ with $Z$ and $\mathcal{I}_z = \{(Z, A)\}$ is destructive since:*

   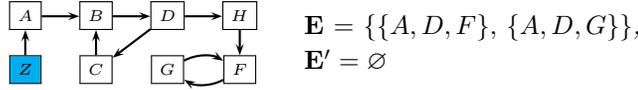

   $\mathbf{E} = \{\{A, D, F\}, \{A, D, G\}\}$,
   $\mathbf{E}' = \varnothing$

2. *Under the preferred (or grounded) semantics, the change $\oplus_i^a$ with $Z$ and $\mathcal{I}_z = \{(Z, A), (B, Z)\}$ is destructive since:*

   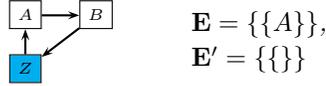

   $\mathbf{E} = \{\{A\}\}$,
   $\mathbf{E}' = \{\{\}\}$

3. *Under the preferred semantics, the change $\oplus_i^a$ with $Z$ and $\mathcal{I}_z = \{(Z, A), (Z, B), (F, Z)\}$ is destructive since:*

   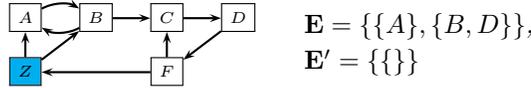

   $\mathbf{E} = \{\{A\}, \{B, D\}\}$,
   $\mathbf{E}' = \{\{\}\}$

So far, the considered changes have an impact on the number of extensions. Now, we are interested in changes which may modify the content of extensions, without modifying the number of extensions. The most interesting situation occurs when each extension of $\mathcal{G}'$ strictly includes one extension of $\mathcal{G}$, the number of extensions being the same. Such changes are called **expansive**.

**Definition 12 (Expansive change)** *The change from $\mathcal{G}$ to $\mathcal{G}'$ is expansive iff $\mathcal{G}$ and $\mathcal{G}'$ have the same number of extensions and each extension of $\mathcal{G}'$ strictly includes an extension of $\mathcal{G}$.*

**Example 6** *Under the preferred (or stable) semantics, the change $\oplus_i^a$ with $Z$ and $\mathcal{I}_z = \{(B, Z)\}$ is expansive since:*

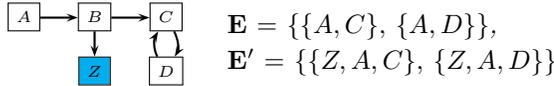

$\mathbf{E} = \{\{A, C\}, \{A, D\}\}$,
$\mathbf{E}' = \{\{Z, A, C\}, \{Z, A, D\}\}$

In the particular case when the set of extensions remains unchanged, the change is called **conservative**.

**Definition 13 (Conservative change)** *The change from $\mathcal{G}$ to $\mathcal{G}'$ is conservative iff $\mathcal{G}$ and $\mathcal{G}'$ have exactly the same extensions, that is $\mathbf{E} = \mathbf{E}'$.*

**Example 7**

1. *Under the preferred semantics, the change $\oplus_i^a$ with $Z$ and $\mathcal{I}_z = \{(B, Z)\}$ is conservative since:*

   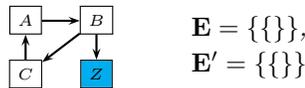

   $\mathbf{E} = \{\{\}\}$,
   $\mathbf{E}' = \{\{\}\}$

2. *Under the preferred semantics, the change $\oplus_i^a$ with $Z$ and $\mathcal{I}_z = \{(A, Z)\}$ is conservative since:*

   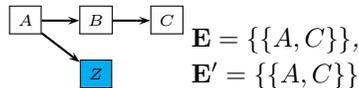

   $\mathbf{E} = \{\{A, C\}\}$,
   $\mathbf{E}' = \{\{A, C\}\}$

3. *Under the preferred semantics, the change $\oplus_i^a$ with $Z$ and $\mathcal{I}_z = \{(A, Z)\}$ is conservative since:*

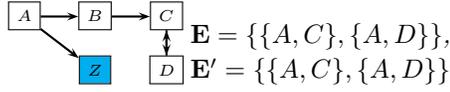

$\mathbf{E} = \{\{A, C\}, \{A, D\}\},$
$\mathbf{E'} = \{\{A, C\}, \{A, D\}\}$

Otherwise, it may happen that $\mathcal{G}$ and $\mathcal{G}'$ have the same number of extensions but some extensions (and sometimes all of them) are altered. This is called an **altering** change.

**Definition 14 (Altering change)** *The change from $\mathcal{G}$ to $\mathcal{G}'$ is altering iff $\mathcal{G}$ and $\mathcal{G}'$ have the same number of extensions and there exists at least one extension $\mathcal{E}_i$ of $\mathcal{G}$ such that $\forall \mathcal{E}_j'$ extension of $\mathcal{G}'$, $\mathcal{E}_i \nsubseteq \mathcal{E}_j'$.*

It is the case for instance when each extension of $\mathcal{G}'$ has a non-empty intersection with (but does not include) an extension of $\mathcal{G}$.

**Example 8**

1. *Under the grounded semantics, the change $\oplus_i^a$ with $Z$ and $\mathcal{I}_z = \{(Z, A)\}$ is altering since:*

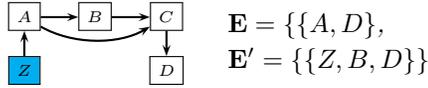

$\mathbf{E} = \{\{A, D\},$
$\mathbf{E'} = \{\{Z, B, D\}\}$

2. *Under the preferred semantics, the change $\oplus_i^a$ with $Z$ and $\mathcal{I}_z = \{(Z, E), (F, Z)\}$ is altering since:*

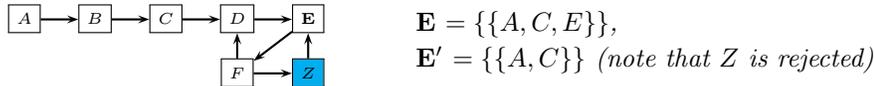

$\mathbf{E} = \{\{A, C, E\}\},$
$\mathbf{E'} = \{\{A, C\}\}$ *(note that $Z$ is rejected)*

The above discussion can be summarized on Table 2. In this table, it can be checked that cells with #i correspond to situations which cannot occur:

**#1 and #2** The only acceptability semantics in which an argumentation framework may have no extension is the stable semantics. However, with the stable semantics, an argumentation framework cannot have an empty extension when its set of arguments is not empty. And, by assumption, the cases #1 and #2 correspond to argumentation frameworks with non-empty sets of arguments (because by assumption either $\mathcal{I}_z \neq \varnothing$ or there exists one interaction $i = (X, Y)$, so there is at least one $X$ in $\mathcal{G}$ and this $X$ and eventually $Z$ belong to $\mathcal{G}'$). So these cases do not occur for any change operation and any acceptability semantics considered in this paper.

Note that the structural properties presented on Table 2 are mutually exclusive (that is a change operation cannot satisfy two of them).

| $\mathbf{E}' =$ <br> $\mathbf{E} =$ | $\varnothing$ | $\{\{\}\}$ | $\{\mathcal{E}'_1\}$ | $\{\mathcal{E}'_1,\ldots,\mathcal{E}'_p\}$ <br> $p \geq 2$ |
|---|---|---|---|---|
| $\varnothing$ | conservative | #1 | decisive | questioning |
| $\{\{\}\}$ | #2 | conservative | | |
| $\{\mathcal{E}_1\}$ | destructive | | conservative <br> expansive <br> altering | questioning |
| $\{\mathcal{E}_1,\ldots,\mathcal{E}_n\}$ <br> $n \geq 2$ | | | decisive | $n < p$: <br> questioning <br> $n > p$: <br> restrictive <br> $n = p$: <br> conservative <br> expansive <br> altering |

With $\mathcal{E}_i \neq \{\}$ and $\mathcal{E}'_i \neq \{\}$. Each cell of the table contains the name of the corresponding property for a change operation.

Table 2: Structural properties of a change operation

## 3.3 Status-Based Properties

In this section, we are interested in the impact of a change operation on the status of some particular arguments.

- First, we are interested in the status of the arguments which were accepted before change. This leads to propose a property called "Monotony", which can be defined for any type of change.

- Another interesting issue concerns the status of the argument which is added in a change. Obviously, it concerns only the change operation $\oplus_i^a$; This leads to propose a property called "Priority to Recency" which only makes sense for one type of change.

### 3.3.1 MONOTONY

Inspired by what has been done in the field of non-monotonic inference, we define a property of monotony for expressing that arguments accepted before change remain accepted after change. Since our aim is to define general properties, we make no assumption about the number of extensions, and we have to consider different cases for acceptance of an argument (credulously or skeptically accepted).

A monotony definition is straightforward under a semantics providing only one extension (such as the grounded semantics, for instance). Following Definition 6, an argument $A$ is accepted (credulously or skeptically) in $\langle \mathbf{A}, \mathbf{R} \rangle$ iff it belongs to the (unique) extension of $\mathcal{G}$. So, in that particular case, monotony means that the extension of $\mathcal{G}$ is included in the extension of $\mathcal{G}'$. When there are several extensions, monotony can take different forms. A credulous form corresponds to the case where each argument credulously accepted from $\mathcal{G}$ is also credulously accepted from $\mathcal{G}'$. A skeptical form corresponds to the case when each argument skeptically accepted from $\mathcal{G}$ is also skeptically accepted from $\mathcal{G}'$. So these ideas lead to the following definition:

**Definition 15 (Monotony)**

- *The change from $\mathcal{G}$ to $\mathcal{G}'$ satisfies* Monotony *iff each extension of $\mathcal{G}$ is included in at least one extension of $\mathcal{G}'$.*

- *The change from $\mathcal{G}$ to $\mathcal{G}'$ satisfies* Credulous Monotony[6] *iff the union of the extensions of $\mathcal{G}$ is included in the union of the extensions of $\mathcal{G}'$.*

- *The change from $\mathcal{G}$ to $\mathcal{G}'$ satisfies* Skeptical Monotony *iff the intersection of the extensions of $\mathcal{G}$ is included in the intersection of the extensions of $\mathcal{G}'$.*

For the change operation $\oplus_i^a$, Examples 2.1, 2.2, 4.3, 6, 7 illustrate the case when the property of Monotony holds; and, again for the change operation $\oplus_i^a$, Examples 2.3, 2.4, 3.1, 3.2, 4.1, 4.2, 5, 8.1, 8.2 illustrate the case when the property of Monotony does not hold[7].

Obviously, Monotony implies Credulous Monotony. However, Monotony does not imply Skeptical Monotony (see Example 4. 3) and Skeptical Monotony does not imply Monotony (see Examples 2.3, 2.4, 3.1, 3.2). Under a semantics providing only one extension, the three notions of Monotony coincide.

The Monotony property is defined at the level of extensions. A similar notion can be defined at the level of arguments:

**Definition 16 (Partial Monotony for an argument)** *Let $X$ be an argument.*
*The change from $\mathcal{G}$ to $\mathcal{G}'$ satisfies* Partial Monotony for $X$ *iff when $X$ belongs to an extension of $\mathcal{G}$, it also belongs to at least one extension of $\mathcal{G}'$.*

It is easy to prove that Monotony (resp. Credulous Monotony) implies Partial Monotony for each argument of $\mathcal{G}$. It is not the case with the property of Skeptical Monotony (see the argument $A$ in Example 2.4).

### 3.3.2 Priority to Recency

The next property concerns the status of the argument which is added in a change. Inspired by what has been done in the field of belief revision (see Alchourrón, Gärdenfors, & Makinson, 1985), and the postulate concerning the priority of the new piece of information, we define a property for expressing that the new argument is accepted after change. This property called **Priority to Recency**[8] makes sense only for the change operation $\oplus_i^a$.

---

6. Credulous Monotony is related to the well-known decision problem of credulous acceptance in argumentation (see Definition 6).
7. In Appendix B, the reader will find some examples illustrating the property of Monotony for the other change operations.
8. This property is not a characteristic postulate in AGM's sense; it has just been inspired by the "Success" postulate proposed by Alchourrón et al. (1985).

**Definition 17 (Priority to Recency)** *The change $\oplus_i^a$ from $\mathcal{G}$ to $\mathcal{G}'$ satisfies* Priority to Recency *iff $\mathcal{G}'$ has at least one extension and the added argument $Z$ belongs to each extension of $\mathcal{G}'$.*

Examples 2.1 to 2.3, 3.1, 4.1, 4.2, 6, 8.1 are examples of change satisfying Priority to Recency. Examples 2.4, 3.2, 4.3, 5, 7, 8.2 are examples of change that do not satisfy Priority to Recency.

### 3.4 Connections between Properties

Some links between structural properties and status-based properties can be established. The following propositions enumerate results that hold for any type of change.

**Proposition 3**

- *A conservative change always satisfies Monotony and Skeptical Monotony.*

- *An expansive change always satisfies Monotony and Skeptical Monotony.*

- *A decisive change which satisfies Monotony also satisfies Skeptical Monotony.*

- *In the particular case of a semantics providing only one extension, a change satisfies Monotony (and Skeptical Monotony) iff it is either decisive, or expansive, or conservative.*

**Proposition 4**

- *A destructive change never satisfies Monotony.*

- *An altering change never satisfies Monotony.*

- *A restrictive change never satisfies Monotony.*

Moreover, in the particular case of the change $\oplus_i^a$, other connections between structural properties and Priority to Recency can be established.

**Proposition 5**

- *A conservative change $\oplus_i^a$ never satisfies Priority to Recency.*

- *A destructive change $\oplus_i^a$ never satisfies Priority to Recency.*

And in the particular case of grounded, stable and preferred semantics, we have:

**Proposition 6** *Under the grounded, stable and preferred semantics, an expansive change $\oplus_i^a$ always satisfies Priority to Recency.*

From the above results and examples given in Sections 3.2 and 3.3, inclusion links between different changes of the type $\oplus_i^a$ are synthesized on Figure 1[9]. Table 3 gives the references of the examples and propositions used for identifying these links.

---

9. The inclusion of "Expansive changes" into the operations that satisfy "Priority to Recency" that is shown in Figure 1, was checked only for the stable, grounded and preferred semantics – see Proposition 6 (hence, it may not hold for other semantics).

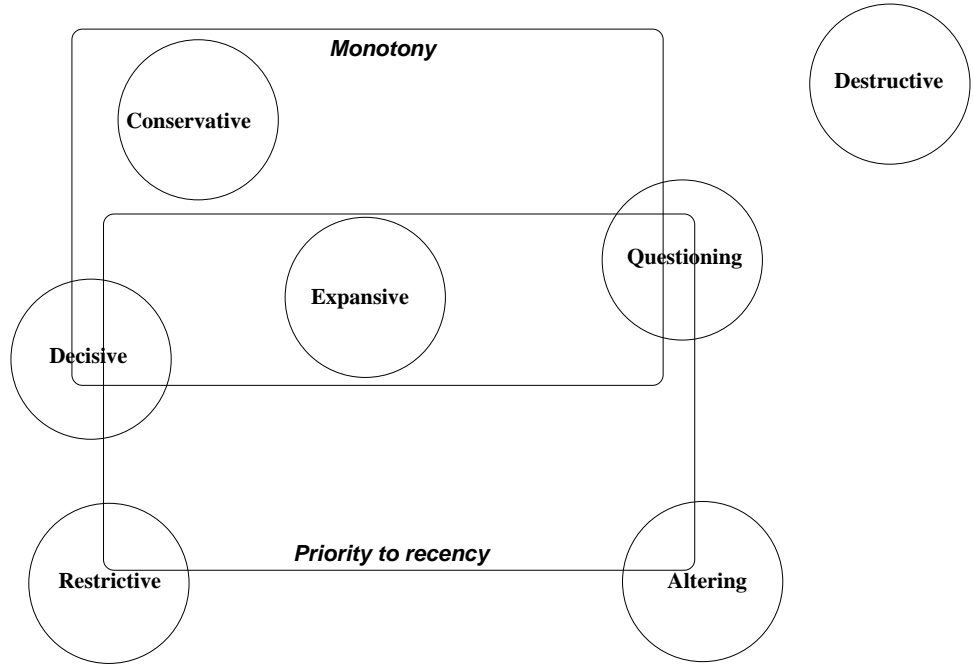

Figure 1: Inclusion links between changes of the type $\oplus_i^a$

| | Priority to Recency | | Monotony | |
|---|---|---|---|---|
| conservative | Never satisfied (Conseq. 5) | | Always satisfied(Conseq. 3) | |
| decisive | May hold (Ex. 2.1 to 2.3) | and not (Ex. 2.4) | May hold (Ex. 2.1) | and not (Ex. 2.3) |
| destructive | Never satisfied (Conseq. 5) | | Never satisfied (Conseq. 4) | |
| expansive | Hold under stable, grounded, preferred sem. (Prop. 6) | | Always (Conseq. 3) | |
| altering | May hold (Ex. 8.1) | and not (Ex. 8.2) | Never (Conseq. 4) | |
| questioning | May hold (Ex. 4.1) | and not (Ex. 4.3) | May hold (Ex. 4.3) | and not (Ex. 4.1) |
| restrictive | May hold (Ex. 3.1) | and not (Ex. 3.2) | Never (Conseq. 4) | |

Table 3: Synthesis about connections between structural and status-based properties of $\oplus_i^a$

## 4. Characterizing Argument Addition under Grounded or Preferred Semantics

In this section, we focus on the change $\oplus_i^a$, *i.e.*, the addition of *exactly* one argument $Z$ that interacts with *at least one* argument belonging to **A**. Indeed, adding an argument which may interact with the existing ones is a very frequently encountered type of change in real-life situations. Besides, this type of change is sufficiently complex to provide a rich analysis of properties and results.

Moreover, we consider the change $\oplus_i^a$ under the grounded and the preferred semantics. We have chosen these two semantics because they are the most well-known traditional semantics for which the existence of extensions is guaranteed.

Our purpose is to identify conditions under which a given property is satisfied for a change operation $\oplus_i^a$. These conditions concern the added argument and the associated interactions, and may depend on the semantics.

Arguably, some properties seem more desirable than others according to the context. For instance, a *decisive* change operation will reduce ignorance, since after the change one and only one extension remains, enabling to determine the status of each argument (which was not always the case before the change). An *expansive* change will raise the number of accepted arguments, which is interesting for achieving a goal of persuasion for instance. A *conservative* change keeps the extensions unchanged, and so is interesting if we want to add an argument without changing our state of knowledge. The properties of Monotony and Priority to Recency are desirable when we focus on some particular arguments, which we want to get in the resulting extensions.

In contrast, a *questioning* or *destructive* operation will increase ignorance, which seems to be less interesting.

An *altering* operation enforces to have a new look at the problem, since nothing is kept from the state before change (the same number of extension remains but they are all different from the previous ones). According to this discussion, we provide:

- sufficient conditions (CS) *under which some interesting properties hold* (e.g., decisive, expansive, conservative, monotonic, satisfying Priority to Recency);

- necessary conditions (CN) for *some undesirable properties* (e.g., questioning, destructive, altering), in order to avoid those properties.

In the following subsections, we consider the change $\oplus_i^a$ with the addition of the argument $Z$ and the interactions $\mathcal{I}_z$, such that:

$$\langle \mathbf{A}, \mathbf{R} \rangle \oplus_i^a (Z, \mathcal{I}_z) = \langle \mathbf{A} \cup \{Z\}, \mathbf{R} \cup \mathcal{I}_z \rangle$$

### 4.1 Argument Addition under the Grounded Semantics

Under the grounded semantics, we have $\mathbf{E} = \{\mathcal{E}\}$ and $\mathbf{E}' = \{\mathcal{E}'\}$.

The following result gives a condition under which a given accepted argument $X$ remains accepted after the change $\oplus_i^a$ (hence Partial Monotony holds for $X$).

**Proposition 7** *Under the grounded semantics, if $X$ belongs to $\mathcal{E}$, and $Z$ does not indirectly attack $X$, then $\oplus_i^a$ satisfies Partial Monotony for $X$ (i.e. $X$ belongs to $\mathcal{E}'$).*

**Example 9** *Under the grounded semantics:*

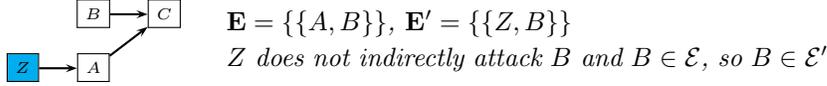

$\mathbf{E} = \{\{A, B\}\}$, $\mathbf{E'} = \{\{Z, B\}\}$

$Z$ *does not indirectly attack* $B$ *and* $B \in \mathcal{E}$, *so* $B \in \mathcal{E'}$

The following result gives a condition under which the change $\oplus_i^a$ satisfies Priority to Recency.

**Proposition 8** *Under the grounded semantics, if* $Z$ *is not attacked by* $\mathcal{G}$, *then* $\oplus_i^a$ *satisfies Priority to Recency (i.e.* $Z$ *belongs to* $\mathcal{E'}$).

**Example 10** *Under the grounded semantics:*

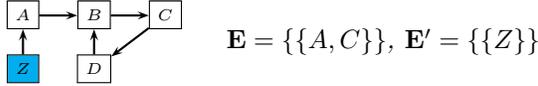

$\mathbf{E} = \{\{A, C\}\}$, $\mathbf{E'} = \{\{Z\}\}$

Let us first study the particular case when $\mathcal{E} = \{\}$.

**Proposition 9** *Under the grounded semantics,*

- *if* $\mathcal{E} = \{\}$ *then the following equivalence holds:* $\mathcal{E'} = \{\}$ *iff* $Z$ *is attacked by* $\mathcal{G}$;

- *moreover, if* $\mathcal{E} = \{\}$ *and* $Z$ *is not attacked by* $\mathcal{G}$, *then* $\mathcal{E'} = \{Z\} \cup \bigcup_{i \geq 1} \mathcal{F}'^i(\{Z\})$.

So, in case $\mathcal{E} = \{\}$, we have:

- Either $Z$ is attacked by $\mathcal{G}$ and then $\mathcal{E'} = \{\}$ (and the change $\oplus_i^a$ is conservative).

- Or $Z$ is not attacked by $\mathcal{G}$ and then $\mathcal{E'}$ contains $Z$ and all the arguments which are indirectly defended by $Z$ (and the change $\oplus_i^a$ is decisive).

As a consequence of Proposition 9, we have:

**Corollary 1** *Under the grounded semantics,*

- *if* $\mathcal{E} = \{\}$ *and* $Z$ *is not attacked by* $\mathcal{G}$, *then the change* $\oplus_i^a$ *is decisive;*

- *if the change* $\oplus_i^a$ *is decisive, then* $Z$ *is not attacked by* $\mathcal{G}$ *and hence* $Z$ *attacks* $\mathcal{G}$.

**Example 11** *Under the grounded semantics, the following change* $\oplus_i^a$ *is decisive:*

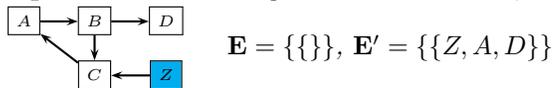

$\mathbf{E} = \{\{\}\}$, $\mathbf{E'} = \{\{Z, A, D\}\}$

Now, we study the particular case when $\mathcal{E} \neq \{\}$.

The following result gives a condition under which the change $\oplus_i^a$ satisfies Monotony.

**Proposition 10** *Under the grounded semantics, if* $\mathcal{E} \neq \{\}$ *and* $Z$ *does not attack* $\mathcal{E}$, *then* $\oplus_i^a$ *satisfies Monotony (i.e.* $\mathcal{E} \subseteq \mathcal{E'}$).

And more precisely, we have two conditions (one for a conservative change $\oplus_i^a$ and another one for an expansive change $\oplus_i^a$):

**Proposition 11** *Under the grounded semantics, if $\mathcal{E} \neq \{\}$ and $Z$ does not attack $\mathcal{E}$, we have:*

- *if $\mathcal{E}$ does not defend $Z$, then $\mathcal{E}' = \mathcal{E}$. (The change $\oplus_i^a$ is conservative).*

- *if $\mathcal{E}$ defends $Z$, then $\mathcal{E}' = \mathcal{E} \cup \{Z\} \cup \bigcup_{i \geq 1} \mathcal{F}'^i(\{Z\})$. Moreover, in that case, if $Z$ does not attack $\mathcal{G}$, $\mathcal{E}'$ reduces to $\mathcal{E} \cup \{Z\}$. (The change $\oplus_i^a$ is expansive).*

**Example 12** *Under the grounded semantics, the following change $\oplus_i^a$ is expansive:*

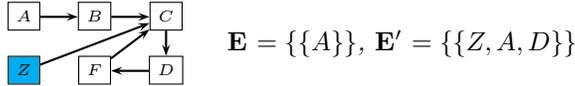   $\mathbf{E} = \{\{A\}\}$, $\mathbf{E}' = \{\{Z, A, D\}\}$

As a consequence of Proposition 11, we have another condition under which the change $\oplus_i^a$ satisfies Priority to Recency:

**Corollary 2** *Under the grounded semantics, if $\mathcal{E} \neq \{\}$, $Z$ does not attack $\mathcal{E}$, and $\mathcal{E}$ defends $Z$, then $\oplus_i^a$ satisfies Priority to Recency (i.e. $Z$ belongs to $\mathcal{E}'$).*

**Example 13** *Under the grounded semantics:*

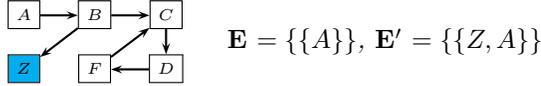   $\mathbf{E} = \{\{A\}\}$, $\mathbf{E}' = \{\{Z, A\}\}$

Note that Corollary 2 does not hold if $\mathcal{E}$ does not defend $Z$.

**Example 14** *Under the grounded semantics:*

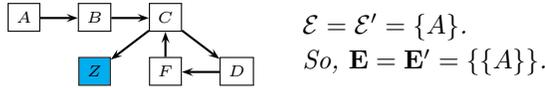   $\mathcal{E} = \mathcal{E}' = \{A\}$.
   *So,* $\mathbf{E} = \mathbf{E}' = \{\{A\}\}$.

Another interesting point is the fact that some properties of the change $\oplus_i^a$ cannot be satisfied under the grounded semantics:

**Proposition 12** *Under the grounded semantics, a change $\oplus_i^a$ is never questioning, nor restrictive.*

The case of a destructive change $\oplus_i^a$ is also interesting because it is sufficient to add an attack against each unattacked argument for obtaining such a change:

**Proposition 13** *Under the grounded semantics, if $\mathcal{E} \neq \{\}$, if $Z$ attacks each unattacked argument $A_i$ of $\mathcal{G}$ and if $Z$ is attacked in $\mathcal{G}'$ then the change $\oplus_i^a$ is destructive; the converse also holds.*

## 4.2 Argument Addition under the Preferred Semantics

Under the preferred semantics, there is always at least one extension. **E** may be reduced to a singleton $\{\mathcal{E}_1\}$ (where $\mathcal{E}_1$ may be empty), or may contain more than one extension $\{\mathcal{E}_1, \ldots, \mathcal{E}_n\}$. Similarly, **E'** may be reduced to a singleton $\{\mathcal{E}_1'\}$ (where $\mathcal{E}_1'$ may be empty), or may contain more than one extension $\{\mathcal{E}_1', \ldots, \mathcal{E}_n'\}$.

The following result gives a condition under which the change $\oplus_i^a$ satisfies Priority to Recency.

**Proposition 14** *Under the preferred semantics, if $Z$ is not attacked by $\mathcal{G}$, then $\oplus_i^a$ satisfies Priority to Recency (i.e. $Z$ belongs to each $\mathcal{E}_i'$).*

**Example 15** *Under the preferred semantics:*

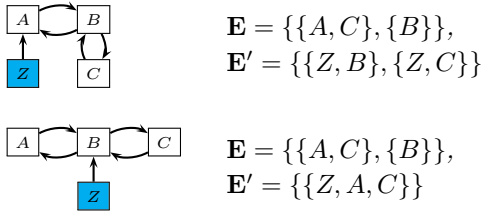

$\mathbf{E} = \{\{A, C\}, \{B\}\},$
$\mathbf{E}' = \{\{Z, B\}, \{Z, C\}\}$

$\mathbf{E} = \{\{A, C\}, \{B\}\},$
$\mathbf{E}' = \{\{Z, A, C\}\}$

The following proposition establishes that admissible sets of $\mathcal{G}$ can be kept in some cases (so, in these cases the change $\oplus_i^a$ can be neither altering, nor restrictive):

**Proposition 15** *Under the preferred semantics,*

- *if $Z$ does not attack $\mathcal{E}_i$, then $\mathcal{E}_i$ remains admissible in $\mathcal{G}'$;*

- *if $Z$ does not attack $\mathcal{E}_i$ and $\mathcal{E}_i$ defends $Z$, then $\mathcal{E}_i \cup \{Z\}$ is admissible in $\mathcal{G}'$.*

**Example 16** *Under the preferred semantics:*

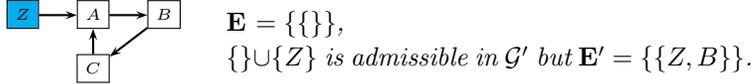

$\mathbf{E} = \{\{\}\},$
*$\{\} \cup \{Z\}$ is admissible in $\mathcal{G}'$ but $\mathbf{E}' = \{\{Z, B\}\}$.*

**Example 12 (continued)** *Under the preferred semantics, $\mathbf{E} = \{\{A\}\}$, $\{A\} \cup \{Z\}$ is admissible in $\mathcal{G}'$, nevertheless, $\mathbf{E}' = \{\{Z, A, D\}\}$.*

Note that other preferred extensions may appear in $\mathcal{G}'$.

**Example 17** *Under the preferred semantics:*

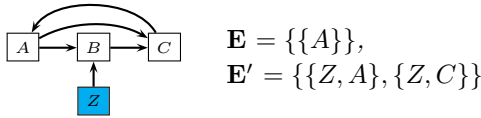

$\mathbf{E} = \{\{A\}\},$
$\mathbf{E}' = \{\{Z, A\}, \{Z, C\}\}$

As a consequence of Proposition 15, we have another condition under which the change $\oplus_i^a$ satisfies Priority to Recency.

**Corollary 3** *Under the preferred semantics, if $Z$ attacks no extension of $\mathcal{G}$, and if each $\mathcal{E}_i$ defends $Z$, then $\oplus_i^a$ satisfies Priority to Recency (i.e. $Z$ belongs to each $\mathcal{E}_i'$).*

**Example 18** *Under the preferred semantics:*

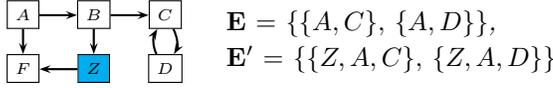

$\mathbf{E} = \{\{A, C\}, \{A, D\}\},$

$\mathbf{E}' = \{\{Z, A, C\}, \{Z, A, D\}\}$

The following result gives a condition under which the change $\oplus_i^a$ is decisive.

**Proposition 16** *Under the preferred semantics, if $\mathbf{E} = \{\{\}\}$ and $Z$ is not attacked by $\mathcal{G}$ and there is no even-length cycle in $\mathcal{G}$ then $\mathbf{E}' = \{\mathcal{E}'\}$ and $Z$ belongs to $\mathcal{E}'$ (so, $\oplus_i^a$ is decisive).*

**Example 11 (continued)** *Under the preferred semantics, $\mathbf{E} = \{\{\}\}$, $\mathbf{E}' = \{\{Z, A, D\}\}$*

Note that, if even-length cycles exist in the graph, the change $\oplus_i^a$ may induce several extensions; this change would be a questioning one:

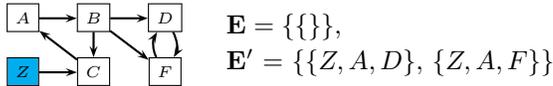

$\mathbf{E} = \{\{\}\},$

$\mathbf{E}' = \{\{Z, A, D\}, \{Z, A, F\}\}$

For this reason, we have considered graphs without even-length cycle in Proposition 16.

The following result gives a necessary condition for $\oplus_i^a$ to be a decisive change (and also a condition for being a conservative change).

**Proposition 17** *Under the preferred semantics, if $Z$ attacks no argument of $\mathcal{G}$ and $\mathbf{E} = \{\{\}\}$, then $\mathbf{E}' = \{\{\}\}$; or equivalently, if $\mathbf{E} = \{\{\}\}$ the change $\oplus_i^a$ by $Z$ is decisive only if $Z$ attacks $\mathcal{G}$.*

The following result relates to the case where there exists a non empty extension in $\mathcal{G}$ and also gives conditions for $\oplus_i^a$ either to be a conservative change, or to be an expansive one.

**Proposition 18** *Under the preferred semantics, if $Z$ attacks no argument of $\mathcal{G}$, and $\mathbf{E} \neq \{\{\}\}$, then for each $i$:*

- *if $\mathcal{E}_i$ defends $Z$, then $\mathcal{E}_i \cup \{Z\}$ is an extension of $\mathcal{G}'$;*

- *if $\mathcal{E}_i$ does not defend $Z$, then $\mathcal{E}_i$ is an extension of $\mathcal{G}'$;*

*moreover, $\mathcal{G}$ and $\mathcal{G}'$ have the same number of extensions.*

**Example 6 (continued)** *Under the preferred semantics, the change $\oplus_i^a$ is expansive:* $\mathbf{E} = \{\{A, C\}, \{A, D\}\}$ *and* $\mathbf{E}' = \{\{Z, A, C\}, \{Z, A, D\}\}$

As a consequence of the previous results, we have a condition under which the change $\oplus_i^a$ satisfies Monotony.

**Proposition 19** *Under the preferred semantics, if $Z$ attacks no extension of $\mathcal{G}$ then the change $\oplus_i^a$ satisfies Monotony.*

In the particular case of a non controversial argumentation framework, we obtain a condition under which the change $\oplus_i^a$ satisfies Skeptical Monotony. The notion of controversial argument has been introduced by Dung, who has proved that an argumentation framework without any controversial argument has nice properties. Roughly speaking, an argument $X$ is controversial if it indirectly attacks and indirectly defends a same argument $Y$.

**Proposition 20** *Under the preferred semantics, assume that $\mathcal{G}$ contains no controversial argument. If $Z$ does not attack $\cap_{i \geq 1} \mathcal{E}_i$, then the change $\oplus_i^a$ satisfies Skeptical Monotony, that is $\cap_{i \geq 1} \mathcal{E}_i \subseteq \cap_{i \geq 1} \mathcal{E}_i'$.*

As under the grounded semantics, there exists a proposition about the destructive change $\oplus_i^a$:

**Proposition 21** *Under the preferred semantics, if $\mathbf{E} \neq \{\{\}\}$, if there is no even-length cycle in $\mathcal{G}'$, if each unattacked argument $A_i$ of $\mathcal{G}$ is attacked in $\mathcal{G}'$ and if $Z$ is attacked in $\mathcal{G}'$ then the change $\oplus_i^a$ is destructive.*

### 4.3 Synthesis of the Results

Tables 4 and 5, display a summary of necessary (CN) or sufficient (CS) conditions for a property to hold for a change $\oplus_i^a$ (in some cases, several CS – resp. CN – may be given denoted by CS, CS', . . . – resp. CN, CN', . . .).

In these tables, $\mathbf{E}, \mathbf{E}', \mathcal{E}, \mathcal{E}', \mathcal{E}_i, \mathcal{E}_j'$ denote respectively the set of extensions before change, after change, the grounded extension before change, after change, a preferred extension before change and after change.

Table 4 concerns the structural properties for a change $\oplus_i^a$.

Table 5 concerns the status-based properties for a change $\oplus_i^a$.

These tables underline the fact that we have been able to identify sufficient conditions (CS) *under which some interesting properties hold* (e.g., decisive, expansive, conservative, monotonic, satisfying Priority to Recency). For the properties of changes that are less desirable such as questioning, destructive, altering, we have focused our search on necessary conditions (CN), allowing us to enunciate sufficient conditions in order *to avoid them*.

## 5. Discussion and Future Works

In this paper, we study change in argumentation. We propose properties to characterize the impact of a change operation on the outcome of an argumentation framework. Then, we focus on a particular type of change: the addition of a new argument that may interact with previously introduced arguments[10]. And we establish conditions under which a given property is satisfied.

The study of change is an important issue in Artificial Intelligence, but it traditionally concerns belief change. When an agent receives a new piece of information, she must adapt her beliefs; this adaptation is not always easy because it may imply to drop some previous knowledge. The seminal work of Alchourrón, Gärdenfors and Makinson (AGM) (1985) has settled a formal framework for reasoning about belief change and introduced the concept of "belief revision" together with two other types of belief change, namely "contraction" and "expansion". Expansion consists only in adding information without checking its consistency with previous beliefs. Contraction is an operation designed for removing information. Revision consists in adding information while preserving consistency. This last operation is the most interesting one since, in belief theory, inconsistency leads to unexploitable information.

---

10. We do not consider knowledge from which arguments and interactions could be built.

| Properties of the change $\bigoplus_i^a$ | Grounded semantics | Preferred semantics |
|---|---|---|
| **Decisive**<br>($\mathbf{E} = \varnothing$ or $\mathbf{E} = \{\{\}\}$ or $|\mathbf{E}| > 2$) and $|\mathbf{E}'| = 1$ and $\mathbf{E}' \neq \{\{\}\}$ | CS and CN: $\mathcal{E} = \{\}$ and $Z$ not attacked. (Prop.9) | CS: $\mathbf{E} = \{\{\}\}$ and $Z$ not attacked and no even-length cycle in $\mathcal{G}$. (Prop.16)<br>if $\mathbf{E} = \{\{\}\}$ CN: $Z$ attacks $\mathcal{G}$. (Prop.17) |
| **Restrictive**<br>$|\mathbf{E}| > |\mathbf{E}'| > 2$ | Never (Prop.12) | CN: $\exists$ an even-length cycle in $\mathcal{G}$ and $Z$ attacks at least one $\mathcal{E}_i$ (Prop.15) |
| **Questioning**<br>$|\mathbf{E}| < |\mathbf{E}'|$ | Never (Prop.12) | CN: $\exists$ an even-length cycle in $\mathcal{G}'$ and $Z$ attacks $\mathcal{G}$ (Prop.17, Prop.18) |
| **Destructive**<br>$\mathbf{E} \neq \varnothing$ and $\mathbf{E} \neq \{\{\}\}$ and ($\mathbf{E}' = \varnothing$ or $\mathbf{E}' = \{\{\}\}$) | CN and CS: $\mathcal{E} \neq \{\}$ and $Z$ attacks each unattacked argt in $\mathcal{G}$ and $Z$ is attacked (Prop.13) | CS: $\mathbf{E} \neq \{\{\}\}$ and $Z$ is attacked and no even-length cycle in $\mathcal{G}'$ and $Z$ attacks each unattacked argt in $\mathcal{G}$ (Prop.21)<br>CN: $\mathbf{E} \neq \{\{\}\}$ and $Z$ is attacked and $\exists$ an odd-length cycle in $\mathcal{G}'$ and $Z$ attacks each unattacked argt in $\mathcal{G}$ (Prop.1.5, Prop.2.2) |
| **Expansive**<br>$|\mathbf{E}| = |\mathbf{E}'|$ and $\forall \mathcal{E}'_j \in \mathbf{E}'$, $\exists \mathcal{E}_i \in \mathbf{E}$, s.t. $\mathcal{E}_i \subset \mathcal{E}'_j$ | CS: $\mathcal{E} \neq \{\}$ and $Z$ does not attack $\mathcal{E}$ and $\mathcal{E}$ defends $Z$ (Prop.11) | CS: $\mathbf{E} \neq \{\{\}\}$ and $Z$ does not attack $\mathcal{G}$ and $\forall i, \mathcal{E}_i$ defends $Z$ (Prop.18) |
| **Conservative**<br>$\mathbf{E} = \mathbf{E}'$ | CS: $\mathcal{E} = \{\}$ and $Z$ attacked by $\mathcal{G}$ (Prop.9)<br>CS': $\mathcal{E} \neq \{\}$ and $Z$ does not attack $\mathcal{E}$ and $\mathcal{E}$ does not defend $Z$ (Prop.11) | CS: $\mathbf{E} = \{\{\}\}$ and $Z$ does not attack $\mathcal{G}$ (Prop.17)<br>CS': $\mathbf{E} \neq \{\{\}\}$ and $Z$ does not attack $\mathcal{G}$ and $\forall i, \mathcal{E}_i$ does not defend $Z$ (Prop.18) |
| **Altering**<br>$|\mathbf{E}| = |\mathbf{E}'|$ and $\exists \mathcal{E}_i \in \mathbf{E}$ s.t. $\forall \mathcal{E}'_j \in \mathbf{E}'$, $\mathcal{E}_i \nsubseteq \mathcal{E}'_j$ | CN: $\mathcal{E} \neq \{\}$ and $Z$ attacks $\mathcal{E}$ (Prop.10) | CN: $\mathbf{E} \neq \{\{\}\}$ and $\exists \mathcal{E}_i$ s.t. $Z$ attacks $\mathcal{E}_i$ (Prop.15) |

Table 4: Synthesis of the necessary and sufficient conditions (CN and CS) for structural properties – Case of $\bigoplus_i^a$

| Properties of the change $\bigoplus_i^a$ | Grounded Semantics | Preferred Semantics |
|---|---|---|
| **Monotony** <br> $\forall \mathcal{E}_i \in \mathbf{E}, \exists \mathcal{E}_j' \in \mathbf{E}'$, s.t. $\mathcal{E}_i \subseteq \mathcal{E}_j'$ | CS: $\mathcal{E} = \{\}$ <br><br> CS′: $\mathcal{E} \neq \{\}$ and $Z$ does not attack $\mathcal{E}$ (Prop.10) | CS: $Z$ does not attack any $\mathcal{E}_i$ (Prop.19) |
| **Priority to Recency** <br> $|\mathbf{E}'| \geq 1$ and $\forall \mathcal{E}_j' \in \mathbf{E}'$, $Z \in \mathcal{E}_j'$ | CS: $Z$ not attacked (Prop.8) <br> CS′: $\mathcal{E} \neq \{\}$, $Z$ does not attack $\mathcal{E}$ and $\mathcal{E}$ defends $Z$ (Prop.11) | CS: $Z$ not attacked (Prop.14), CS′: $\forall \mathcal{E}_i \in \mathbf{E}$, $Z$ does not attack $\mathcal{E}_i$ and $\mathcal{E}_i$ defends $Z$ (Corol.3) |
| **Partial Monotony for X** <br> If $\exists \mathcal{E}_i \in \mathbf{E}$ s.t. $X \in \mathcal{E}_i$, then $\exists \mathcal{E}_j' \in \mathbf{E}'$ s.t. $X \in \mathcal{E}_j'$ | CS: $X \in \mathcal{E}$ and $Z$ does not indirectly attack $X$ (Prop.7) | *cf.* Monotony (because, $\forall X$, Partial Monotony for $X$ is implied by Monotony) |
| **Skeptical Monotony** <br> $\cap_{i \geq 1} \mathcal{E}_i \subseteq \cap_{j \geq 1} \mathcal{E}_j'$ | *cf.* Monotony (because, for the grounded semantics, Skeptical Monotony is Monotony) | CS: no controversial argt in $\mathcal{G}$ and $Z$ does not attack $\cap_{i \geq 1} \mathcal{E}_i$ (Prop.20) |

Table 5: Synthesis of the necessary and sufficient conditions (CN and CS) for status-based properties – Case of $\bigoplus_i^a$

Although the change operations defined in Section 3 could be thought of as being related to the AGM theory[11], the comparison is not appropriate because of two main reasons:

- The basic underlying formalism is different: in standard belief revision, logical formulae are used for knowledge representation whereas, in this paper, an argumentation framework represents the current knowledge. In the first case, the outcome is a new set of logical formulae, whereas, in the second case, the outcome is a new argumentation framework which induces a new set of extensions, each extension being a set of arguments.

- Revision is a task in knowledge representation which is strongly related to concepts such as inference and consistency. The postulates for standard belief revision (AGM) are built on the consistency notion, since revision aims at incorporating a new piece of information while preserving consistency. However, in the framework of argumentation, the notion of consistency has not a clear and standard accepted meaning (even if some authors propose to take into account a kind of "degree of inconsistency" in the argumentation context as in the works of Matt & Toni, 2008; Besnard & Hunter, 2008).

---

11. Note that other important cognitive tasks linked to belief change theory have already been studied in the field of argumentation, see for instance the work on merging of Coste-Marquis, Devred, Konieczny, Lagasquie-Schiex, and Marquis (2007).

Moreover, "revision" has also been studied in the framework of non-monotonic theories (Witteveen & van der Hoek, 1997) and argumentation theory is linked to non-monotony, but postulates for non-monotonic theories are also based on consistency and inference notions that are not explicitly present in an abstract argumentation system. So, these postulates are not suited for our problem. Some of the belief revision postulates can be restated (this is the case for the property called "Priority to Recency" which has been inspired by the AGM "Success postulate" ), but other principles must be proposed (for instance, we have identified a property called "Monotony" which checks a kind of preservation of the existing extensions by the change process).

Our work is an extension of a previous work (Cayrol et al., 2008) which presented a preliminary step towards a formal characterization of the notion of change in argumentation frameworks. In the work of Cayrol et al. (2008), a change was defined as the addition of one argument and *only one* interaction and we studied only the structural properties and Priority of Recency (called "classicity" by Cayrol et al., 2008). In the new version of this work, proposed in the current paper, we are further taking into account the addition of *several* interactions (so some properties given by Cayrol et al., 2008 do not hold here) and defining new properties around the notion of Monotony. We also look further into the connections between all the proposed properties and into the conditions (necessary or sufficient) for obtaining or avoiding these properties.

There are many other approaches that deal with adding new pieces of information within an argumentation system. The point of view adopted in this family of works is different from ours because of the status of the new piece of information that is added. For instance, Wassermann (1999), as well as Falappa, García, and Simari (2004) and Paglieri and Castelfranchi (2005), define under which conditions, expressed in terms of arguments, unjustified beliefs should become accepted. Pollock and Gillies's (2000) approach studies the properties of knowledge revision under the argumentation point of view, *i.e.*, the problem is to generate a knowledge base in which each piece of information is justified by "good" arguments. The same kind of problem is studied by Amgoud and Vesic (2009) in the context of argument-based decision. Argument-based decision takes as input a set of options, a set of arguments and a defeat relation among them, and returns a status for each option together with a total preorder on the set of options. These authors study under which conditions an option may change its status when a new argument is received and under which conditions this new argument is useless.

Recently, Rotstein, Moguillansky, Falappa, García, and Simari (2008a) have proposed a warrant-prioritized revision operation, which consists in adding an argument to a theory in such a way that this argument is warranted afterwards. Even if the underlying ideas are similar, this work differs from our approach in at least two points:

- First, in the work of Rotstein et al. (2008a), arguments are given a structure through the sub-argument relation, and properties such as minimality, consistency and atom-icity. And the definition of warranted arguments relies upon an evaluation of argumentation lines. In contrast, our approach remains at the most abstract level, and our sets of accepted arguments are computed with the well-known extension-based semantics.

- Secondly, the warrant-prioritized argument revision is designed in order to satisfy the AGM Postulate, corresponding to our property of "Priority to recency", since the added argument must be warranted in the revised theory. Our work follows another direction. We propose an extensive theoretical study of the impact of an addition on the outcome of an abstract argumentation framework, which enables us to define several properties for a change operation.

Concerning the more general question of handling dynamics in argumentation, our proposal can be related to recent works of Boella et al. (2009a, 2009b), and of Rotstein et al. (2008b):

- The work of Boella et al. (2009a, 2009b) studies how the extensions of an argumentation system remain unchanged when the set of arguments or the attacks between them are changed.

  The four types of change we have proposed in Definition 7 have been introduced under different names, respectively attack refinement, attack abstraction, argument refinement and argument abstraction. However, only the operations of attack refinement, attack abstraction and argument abstraction have been studied, and from a more restrictive point of view:

  - Boella et al. (2009a, 2009b) consider only the case when the semantics provides exactly one extension.
  - The principles which are defined correspond to conditions under which a change is conservative, in our terminology. No other property is considered.

  As we focus on the addition of an argument and its interactions, the work of Boella et al. (2009a, 2009b) can be viewed as complementary to our work.

- Rotstein et al. (2008b) introduce the notion of dynamics by considering arguments built from evidence. Evidence is used to determine whether an argument is active (*i.e.* can be used to draw inferences) or inactive. The question addressed by Rotstein et al. (2008b) is: "How a variation of the set of evidence affects the nature of arguments (active or not)?". This question cannot be handled at a pure abstract level and concerns internal dynamics. By contrast, we remain at an abstract level: we are interested by the impact of a change of the abstract framework on the outcome of that framework.

A promising application of our work could be to design dialogue strategies. Indeed, a dialogue may be defined as an exchange (called move) of arguments between two or more, human or artificial, agents under a given protocol. The protocol is a program that defines the set of allowed moves at each step of the dialogue. Each agent has its own aim and may develop its own strategy. Most of the works about dialogue strategies consider that a strategy selects exactly one move (the move which must be done next). For instance, Bench-Capon (1998) proposes a selection strategy (for agents) leading to more cooperative dialogues. Other approaches study strategies in the context of persuasion dialogues, where two agents argue in order to persuade each other that a given initial argument is true (or false according to the agent opinion). In that case, a strategy helps to choose which argument must be defeated in order that the initial argument should be accepted (or rejected). Amgoud and Maudet (2002) have proposed heuristics that select the less attackable arguments in a persuasion dialogue. In a similar way, Riveret, Prakken, Rotolo, and Sartor (2008) have proposed an optimal strategy in order to win a debate based on the probability of success of the argument and on the cost of this argument for the agent. Hunter (2004), with a more global approach, has defined a strategy which builds an optimal subtree of arguments maximizing the resonance with the agent goals and minimizing their cost.

Our approach takes another point of view. We do not define any protocol and we do not restrict to a dialogue type. Given a set of arguments which may interact, we are interested in the outcome of the argumentation system, that is the set of extensions under a given semantics. In other words, we study the impact of the addition of an argument with respect to two points of view: first, the structural modification induced on the set of extensions, second, the impact on the acceptability of arguments. Although the concern of acceptability evolution looks similar to the aim of the existing dialogue approaches presented above, our proposal is more general, since in our work, we are not interested in finding strategies in order to make accepted a precise argument but we are rather interested in establishing general conditions for the preservation of acceptability. For instance, under the grounded and preferred semantics, we provide a sufficient condition for maintaining an argument accepted after the arrival of a new one (Monotony property) and a sufficient condition for a new argument to be accepted (Priority to Recency).

The structural point of view of our analysis is completely original with respect to the existing literature. Indeed, we analyze the impact of a new argument on the set of extensions and this amounts to consider the addition of an argument as an operation performed in order to modify the form of the change outcome (by doing an expansive change, or a decisive change for instance). The work reported in this paper enables us to choose the right way of changing (which argument must be affected by the change, with which kind of interaction) in order to obtain the new outcome. This is why we plan to focus more on strategies for directing a dialogue (*i.e.*, to be integrated in a protocol) than on strategies for taking part in it (*i.e.*, concerning an agent). For instance, if a dialogue arbitrator wants the debate to be more open then she should rather force the next speaker to use arguments appropriate for an expansive change. If she wants the debate to be more focused then only arguments appropriate for a restrictive (and even decisive) change should be accepted.

There are several directions of further research:

1. We plan to study the other change operations defined in this paper, corresponding to the removal of one argument with its interactions and to the addition or the removal of an interaction (for instance, by exploiting properties of symmetry between all the change operations).

2. We would like to generalize our change operations to the case of the addition or the removal of a subgraph of arguments (which would be a kind of "iterated change").

3. We think that the decisive property is a desirable property for a change operation. So, we intend to investigate the question "How to make the *minimal change*[12] to a given argumentation framework so that it has a unique non-empty extension?".

## Acknowledgments

We would like to thank the reviewers for their help and their very interesting suggestions.

## Appendix A. Proofs

**Lemma 1**

- If $\nexists X \in \mathcal{G}$ s.t. $(Z, X) \in \mathcal{I}_z$, the change operation $\oplus_i^a$ introduces no new cycle in $\mathcal{G}'$.

- If $\nexists X \in \mathcal{G}$ s.t. $(X, Z) \in \mathcal{I}_z$, the change operation $\oplus_i^a$ introduces no new cycle in $\mathcal{G}'$.

In other words, if $Z$ does not attack any argument of $\mathcal{G}$, or if $Z$ is not attacked by $\mathcal{G}$, the change operation $\oplus_i^a$ introduces no new cycle in $\mathcal{G}'$.

**Proof of Lemma 1:** If follows directly from the fact that only one argument is added. □

## Proofs Related to Section 3.4 (Connections between Properties)

**Proof of Proposition 3:** It follows directly from the definitions of the properties (Definitions 8, 12, 13, 15). □

**Proof of Proposition 4:** It follows directly from the definitions of the properties (Definitions 11, 9, 14, 15). □

**Proof of Proposition 5:** It follows directly from the definitions of the properties (Definitions 11, 13, 17). □

**Proof of Proposition 6:**

**Grounded semantics:** Let us show that if $\mathcal{E} \subsetneq \mathcal{E}'$ then $Z \in \mathcal{E}'$. Assume that $\mathcal{E} \subsetneq \mathcal{E}'$ and that $Z \notin \mathcal{E}'$. We are going to prove that $\mathcal{E}' \subseteq \mathcal{E}$ (which is in contradiction with the assumption $\mathcal{E} \subsetneq \mathcal{E}'$), by proving that $\mathcal{F}'^i(\{\}) \subseteq \mathcal{F}^i(\{\})$, by induction on $i \geq 1$.

- Basic case ($i = 1$): $Z \notin \mathcal{E}'$ so $Z$ is attacked by $\mathcal{G}$. Thus, if $X \in \mathcal{F}'(\{\})$ then $X$ is in $\mathcal{G}$ and by definition $X$ is unattacked in $\mathcal{G}'$. Then $X$ is also unattacked in $\mathcal{G}$. So, $\mathcal{F}'(\{\}) \subseteq \mathcal{F}(\{\})$.

- Induction hypothesis (for $1 \leq i \leq p$, $\mathcal{F}'^i(\{\}) \subseteq \mathcal{F}^i(\{\})$):

---

12. In terms of number of edges to add or to remove and/or in terms of number of arguments to add or to remove.

– Let us first show that for any subset of arguments $S$ in $\mathcal{G}$ such that $\mathcal{F}'(S) \subseteq \mathcal{G}$, we have $\mathcal{F}'(S) \subseteq \mathcal{F}(S)$: Let $X \in \mathcal{F}'(S)$, it means that $S$ defends $X$ in $\mathcal{G}'$ and $X \in \mathcal{G}$. If there exists $Y$ in $\mathcal{G}$ which attacks $X$ in $\mathcal{G}$, then $Y$ also attacks $X$ in $\mathcal{G}'$. And as $S$ defends $X$ in $\mathcal{G}'$, then $S$ attacks $Y$. So $S$ also defends $X$ in $\mathcal{G}$. So $\mathcal{F}'(S) \subseteq \mathcal{F}(S)$.

– Let us compute $\mathcal{F}'^{p+1}(\{\}) = \mathcal{F}'(\mathcal{F}'^p(\{\}))$. By induction hypothesis, $\mathcal{F}'^p(\{\}) \subseteq \mathcal{F}^p(\{\})$. As $\mathcal{F}'$ is monotonic, we have $\mathcal{F}'^{p+1}(\{\}) \subseteq \mathcal{F}'(\mathcal{F}^p(\{\}))$. Now let $S$ denote the set $\mathcal{F}^p(\{\})$, $S \subseteq \mathcal{E}$ so $S \subseteq \mathcal{G}$. We have assumed that $\mathcal{E} \subsetneq \mathcal{E}'$ and $Z \notin \mathcal{E}'$ so $S \subseteq \mathcal{E}' \subseteq \mathcal{G}$. Then, as $\mathcal{F}'$ is monotonic, we have $\mathcal{F}'(S) \subseteq \mathcal{F}'(\mathcal{E}') = \mathcal{E}' \subseteq \mathcal{G}$ Due to the previous point, we can conclude that $\mathcal{F}'(S) \subseteq \mathcal{F}(S)$. Then, we obtain $\mathcal{F}'^{p+1}(\{\}) \subseteq \mathcal{F}(\mathcal{F}^p(\{\})) = \mathcal{F}^{p+1}(\{\})$.

**Preferred semantics:** Given an expansive change of $\mathcal{G}$ by $Z$ and $\mathcal{I}_z$, let us suppose that there exists an extension $\mathcal{E}'_j$ of $\mathcal{G}'$ that does not contain $Z$. Then this extension is included in $\mathcal{G}$. As the change is expansive, there exists an extension $\mathcal{E}_i$ of $\mathcal{G}$ strictly included in $\mathcal{E}'_j$. $\mathcal{E}_i$ is a maximal admissible set for inclusion. Since the inclusion of $\mathcal{E}_i$ inside $\mathcal{E}'_j$ is strict, therefore $\mathcal{E}'_j$ is not admissible in $\mathcal{G}$. $\mathcal{E}'_j$ being an extension of $\mathcal{G}'$, it has no conflict, hence $\mathcal{E}'_j$ does not defend each of its elements in $\mathcal{G}$. It exists $X \in \mathcal{E}'_j$ that is attacked by $Y$ in $\mathcal{G}$ (and thus in $\mathcal{G}'$) and that is not defended by $\mathcal{E}'_j$ in $\mathcal{G}$. This means that $\mathcal{E}'_j$ does not attack $Y$. But, since $\mathcal{E}'_j$ is included in $\mathcal{G}$ it can not attack $Y$ in $\mathcal{G}'$. If there is no edge between an element of $\mathcal{E}'_j$ and $Y$ in $\mathcal{G}$, there is neither such an edge in $\mathcal{G}'$. (Note that $Y$ can be attacked by $Z$ but $Z$ is not in $\mathcal{E}'_j$)

**Stable semantics:** Assume that there exists an extension $\mathcal{E}'_j$ of $\mathcal{G}'$ that does not contain $Z$. As the change is expansive, there exists an extension $\mathcal{E}_i$ of $\mathcal{G}$ strictly included in $\mathcal{E}'_j$. Since the inclusion is strict, there exists $Y$ in $\mathcal{E}'_j$, which does not belong to $\mathcal{E}_i$. And as we have assumed that $\mathcal{E}'_j$ does not contain $Z$, $Y$ is in $\mathcal{G}$. $\mathcal{E}_i$ is a stable extension of $\mathcal{G}$, so $\mathcal{E}_i$ attacks $Y$. Then, as $\mathcal{E}_i$ is included in $\mathcal{E}'_j$ there is a conflict in $\mathcal{E}'_j$, which contradicts the fact that $\mathcal{E}'_j$ is a stable extension.

$\square$

## Proofs Related to Section 4.1 (Under the Grounded Semantics)

**Proof of Proposition 7:** $\mathcal{E}$ is the grounded extension of $\mathcal{G}$. Due to the fact that **R** is finite, we have $\mathcal{E} = \cup_{i \geq 1} \mathcal{F}^i(\{\})$. We prove by induction on $i \geq 1$ that if $X$ belongs to $\mathcal{F}^i(\{\})$ and $Z$ does not indirectly attack $X$, then $X$ belongs to $\mathcal{F}'^i(\{\})$.

- Basic case ($i = 1$): If $X \in \mathcal{F}(\{\})$ then $X$ is not attacked in $\mathcal{G}$. Since $Z$ does not attack $X$, then $X$ remains unattacked in $\mathcal{G}'$ and so belongs to $\mathcal{F}'(\{\})$.

- Induction hypothesis (for $1 \leq i \leq p$, the proposition holds): Let $X \in \mathcal{F}^{p+1}(\{\})$. We have to prove that $X \in \mathcal{F}'^{p+1}(\{\})(= \mathcal{F}'(\mathcal{F}'^p(\{\})))$. Assume that $X$ is attacked by $Y$ in $\mathcal{G}'$. As $Z$ does not attack $X$, $Y$ is in $\mathcal{G}$. As $X \in \mathcal{F}^{p+1}(\{\}) = \mathcal{F}(\mathcal{F}^p(\{\}))$, $\mathcal{F}^p(\{\})$ defends $X$ by attacking $Y$. So there exists $W \in \mathcal{F}^p(\{\})$ which attacks $Y$, which in turn attacks $X$. As $Z$ does not indirectly attack $X$, we are sure that $Z$ does not indirectly attack $W$. Using the induction hypothesis for $W$, we have $W \in \mathcal{F}'^p(\{\})$. So, we have proved that $\mathcal{F}'^p(\{\})$ defends $X$ in $\mathcal{G}'$ and so $X \in \mathcal{F}'^{p+1}(\{\})$.

$\square$

**Proof of Proposition 8:** If $Z$ is not attacked by $\mathcal{G}$, then $Z$ is not attacked in $\mathcal{G}'$. So, due to Proposition 1.5, the grounded extension of $\mathcal{G}'$ contains $Z$. $\square$

**Proof of Proposition 9:**

- If $\mathcal{E} = \{\}$ then each argument of $\mathcal{G}$ is attacked. If $Z$ is attacked by $\mathcal{G}$, then each argument of $\mathcal{G}'$ is attacked and due to Proposition 1.5 and Proposition 1.6, $\mathcal{E}' = \{\}$. If $Z$ is not attacked by $\mathcal{G}$, $Z$ is not attacked in $\mathcal{G}'$, then $Z$ belongs to $\mathcal{E}'$, which is not empty.

- If $\mathcal{E} = \{\}$ and $Z$ is not attacked by $\mathcal{G}$, $Z \in \mathcal{E}'$. As $\mathcal{F}'$ is monotonic, and $\mathcal{E}'$ is a fixed point of $\mathcal{F}'$, we have $\mathcal{F}'^i(\{Z\}) \subseteq \mathcal{E}'$ for each $i \geq 1$, and then $\{Z\} \cup \bigcup_{i \geq 1} \mathcal{F}'^i(\{Z\}) \subseteq \mathcal{E}'$. Let $\mathcal{S}$ denote $\{Z\} \cup \bigcup_{i \geq 1} \mathcal{F}'^i(\{Z\})$. Now, we have to prove that $\mathcal{E}' \subseteq \mathcal{S}$. As $\mathcal{E}'$ is the least fixed point of $\mathcal{F}'$, it is sufficient to prove that $\mathcal{S}$ is a fixed point of $\mathcal{F}'$. Obviously, $\mathcal{F}'(\mathcal{S}) = \{X \in \mathcal{G}'$ s.t. $X$ is not attacked$\} \cup \bigcup_{i \geq 1} \mathcal{F}'^i(\{Z\})$. Since $\mathcal{E} = \{\}$, $\{X \in \mathcal{G}'$ s.t. $X$ is not attacked $\} = \{Z\}$, so $\mathcal{F}'(\mathcal{S}) = \mathcal{S}$ and $\mathcal{S}$ is a fixed point of $\mathcal{F}'$. We have proved that $\mathcal{E}' = \{Z\} \cup \bigcup_{i \geq 1} \mathcal{F}'^i(\{Z\})$.

$\square$

**Proof of Corollary 1:**

- It follows directly from Proposition 9. Due to Definition 8, under the grounded semantics, the change is decisive when $\mathcal{E} = \{\}$ and $\mathcal{E}' \neq \{\}$.

- And as $Z$ interacts with $\mathcal{G}$, if $Z$ is not attacked by $\mathcal{G}$, $Z$ must attack $\mathcal{G}$.

$\square$

**Proof of Proposition 10:** Due to the fact that **R** is finite, we have $\mathcal{E} = \cup_{i \geq 1} \mathcal{F}^i(\{\})$ and $\mathcal{E}' = \cup_{i \geq 1} \mathcal{F}'^i(\{\})$. We prove by induction on $i \geq 1$ that $\mathcal{F}^i(\{\}) \subseteq \mathcal{F}'^i(\{\})$.

- Basic case ($i = 1$): If $Y \in \mathcal{F}(\{\})$ then $Y$ is not attacked in $\mathcal{G}$ and due to the fact that $Z$ does not attack $\mathcal{E}$, $Y$ is not attacked in $\mathcal{G}'$ and $Y \in \mathcal{F}'(\{\})$.

- Induction hypothesis (for $1 \leq i \leq p$, $\mathcal{F}^i(\{\}) \subseteq \mathcal{F}'^i(\{\})$): let $\mathcal{S} = \mathcal{F}^p(\{\})$ and $\mathcal{S}' = \mathcal{F}'^p(\{\})$.

  First, we prove that $\mathcal{F}(\mathcal{S}) \subseteq \mathcal{F}'(\mathcal{S})$. Let $Y \in \mathcal{F}(\mathcal{S})$. Obviously, $\mathcal{F}(\mathcal{S}) \subseteq \mathcal{S}$. So $Y \in \mathcal{E}$ and $Z$ does not attack $Y$ since $Z$ does not attack $\mathcal{E}$. So, if $Y$ is attacked by $A$ in $\mathcal{G}'$ then $Y$ is attacked by $A$ in $\mathcal{G}$. As $Y \in \mathcal{F}(\mathcal{S})$, $\mathcal{S}$ defends $Y$ by attacking $A$. And so $\mathcal{S}$ defends $Y$ in $\mathcal{G}'$, that is $Y \in \mathcal{F}'(\mathcal{S})$.

  Using the induction hypothesis, we have $\mathcal{S} \subseteq \mathcal{S}'$. Moreover, by definition $\mathcal{F}'$ is monotonic. So $\mathcal{F}(\mathcal{S}) = \mathcal{F}^{p+1}(\{\}) \subseteq \mathcal{F}'(\mathcal{S}) \subseteq \mathcal{F}'(\mathcal{S}') = \mathcal{F}'^{p+1}(\{\})$. So, $\mathcal{E} \subseteq \mathcal{E}'$.

$\square$

**Proof of Proposition 11:** $\mathcal{E} \neq \{\}$ and $Z$ does not attack $\mathcal{E}$. Let us first notice that (1) If $Y \in \mathcal{F}'(\mathcal{E})$ and $Y \in \mathcal{G}$, then $Y \in \mathcal{F}(\mathcal{E}) = \mathcal{E}$. Indeed, $Y \in \mathcal{F}'(\mathcal{E})$ means that $\mathcal{E}$ defends $Y$ in $\mathcal{G}'$. So, if $Y \in \mathcal{G}$, $\mathcal{E}$ also defends $Y$ in $\mathcal{G}$, *i.e.*, $Y \in \mathcal{F}(\mathcal{E}) = \mathcal{E}$.

- Due to Proposition 10, we have $\mathcal{E} \subseteq \mathcal{E}'$. So, we just have to prove that if $\mathcal{E}$ does not defend $Z$, then $\mathcal{E}' \subseteq \mathcal{E}$. Indeed, we will prove that $\mathcal{F}'(\mathcal{E}) = \mathcal{E}$. Then, by definition of $\mathcal{E}'$ (least fixed point), it will follow that $\mathcal{E}' \subseteq \mathcal{E}$. Let $Y \in \mathcal{F}'(\mathcal{E})$, as $\mathcal{E}$ does not defend $Z$, hence $Y \in \mathcal{G}$, according to (1), $Y \in \mathcal{F}(\mathcal{E}) = \mathcal{E}$. Conversely, let $Y \in \mathcal{E} = \mathcal{F}(\mathcal{E})$, and let $A$ be an argument that attacks $Y$ in $\mathcal{G}'$. As $Z$ does not attack $\mathcal{E}$, $A \neq Z$, so $A \in \mathcal{G}$, and $\mathcal{E}$ defends $Y$ by attacking $A$. So, $\mathcal{E}$ defends $Y$ in $\mathcal{G}'$ and $Y \in \mathcal{F}'(\mathcal{E})$.

- First, we prove that if $\mathcal{E}$ defends $Z$ then $\mathcal{F}'(\mathcal{E}) = \mathcal{E} \cup \{Z\}$. Due to (1), if $Y \in \mathcal{F}'(\mathcal{E})$ and $Y \in \mathcal{G}$, then $Y \in \mathcal{F}(\mathcal{E}) = \mathcal{E}$. Now, if $\mathcal{E}$ defends $Z$, we have also $Z \in \mathcal{F}'(\mathcal{E})$. So, $\mathcal{F}'(\mathcal{E}) \subseteq \mathcal{E} \cup \{Z\}$. Conversely, let $Y \in \mathcal{F}(\mathcal{E}) = \mathcal{E}$. $\mathcal{E}$ defends $Y$ in $\mathcal{G}$. As $Z$ does not attack $\mathcal{E}$, $Z$ does not attack $Y$, and then $\mathcal{E}$ also defends $Y$ in $\mathcal{G}'$, so $Y \in \mathcal{F}'(\mathcal{E})$. And $Z \in \mathcal{F}'(\mathcal{E})$, so $\mathcal{E} \cup \{Z\} \subseteq \mathcal{F}'(\mathcal{E})$.
  In the particular case when $Z$ does not attack $\mathcal{G}$, $Z$ cannot defend any argument. So, $\mathcal{F}'(\mathcal{E} \cup \{Z\}) = \mathcal{F}'(\mathcal{E})$ and then $\mathcal{F}'(\mathcal{E} \cup \{Z\}) = \mathcal{E} \cup \{Z\}$. That means that $\mathcal{E} \cup \{Z\}$ is a fixed point of $\mathcal{F}'$, and by definition of $\mathcal{E}'$, we have $\mathcal{E}' \subseteq \mathcal{E} \cup \{Z\}$. Due to Proposition 10, we have $\mathcal{E} \subseteq \mathcal{E}'$. So, we have also $\mathcal{E} \cup \{Z\} \subseteq \mathcal{E}'$. Finally, $\mathcal{E}'$ reduces to $\mathcal{E} \cup \{Z\}$.

In the general case, $Z$ attacks $\mathcal{G}$. Let $\mathcal{S}$ denote $\mathcal{E} \cup \{Z\} \cup \bigcup_{i \geq 1} \mathcal{F}'^i(\{Z\})$. We will prove that $\mathcal{E}' = \mathcal{S}$. Obviously, we have $\mathcal{S} \subseteq \mathcal{E}'$ since $\mathcal{E} \cup \{Z\} = \mathcal{F}'(\mathcal{E}) \subseteq \mathcal{E}'$, and $\mathcal{E}'$ contains $\bigcup_{i \geq 1} \mathcal{F}'^i(\mathcal{E}')$, hence contains $\bigcup_{i \geq 1} \mathcal{F}'^i(\{Z\})$, since $\mathcal{F}'$ is monotonic. Conversely, we will prove that $\mathcal{S}$ is a fixed point of $\mathcal{F}'$ and by definition of $\mathcal{E}'$ (least fixed point), it will follow that $\mathcal{E}' \subseteq \mathcal{S}$.

Since $\mathcal{F}'$ is monotonic, we have $\mathcal{F}'(\mathcal{E}) \subseteq \mathcal{F}'(\mathcal{S})$, $\mathcal{F}'(\{Z\}) \subseteq \mathcal{F}'(\mathcal{S})$ and for each $i \geq 2$ $\mathcal{F}'^i(\{Z\}) \subseteq \mathcal{F}'(\mathcal{S})$. So, as $\mathcal{F}'(\mathcal{E}) = \mathcal{E} \cup \{Z\}$, we have $\mathcal{S} \subseteq \mathcal{F}'(\mathcal{S})$. Conversely, let $Y \in \mathcal{F}'(\mathcal{S})$ and assume that $Y \notin \mathcal{E} \cup \{Z\} = \mathcal{F}'(\mathcal{E})$. Then, there exists an attacker $A$ of $Y$ and $A$ is not attacked by $\mathcal{E}$. As $Y \in \mathcal{F}'(\mathcal{S})$, $\mathcal{S}$ must attack $A$. So $\{Z\} \cup \bigcup_{i \geq 1} \mathcal{F}'^i(\{Z\})$ attacks $A$, which means that $Y \in \bigcup_{i \geq 1} \mathcal{F}'^i(\{Z\})$. So, we have proved that if $Y \in \mathcal{F}'(\mathcal{S})$, either $Y \in \mathcal{E} \cup \{Z\}$ or $Y \in \bigcup_{i \geq 1} \mathcal{F}'^i(\{Z\})$, that is $Y \in \mathcal{S}$.

<div align="right">□</div>

**Proof of Corollary 2:** It is a direct consequence of Proposition 11.  □

**Proof of Proposition 12:** It is a direct consequence of the definitions: the restrictive and questioning changes need a number of extensions strictly greater than one, and there exists only one grounded extension.  □

**Proof of Proposition 13:** $\mathcal{E} \neq \{\}$, so there are unattacked arguments denoted by $A_i$ in $\mathcal{G}$. $\forall A_i$, $A_i$ is attacked in $\mathcal{G}'$ and $Z$ is attacked in $\mathcal{G}'$. So there is no unattacked argument in $\mathcal{G}'$, so $\forall i \geq 1$ $\mathcal{F}'^i(\{\}) = \{\}$ and $\mathcal{E}' = \{\}$. So the change is destructive.

Conversely, if the change is destructive, by definition we have $\mathcal{E} \neq \{\}$ and $\mathcal{E}' = \{\}$. Then, due to Proposition 1.5, there is no unattacked argument in $\mathcal{G}'$. So, $Z$ is attacked and each $A_i$ (unattacked argument in $\mathcal{G}$) is also attacked in $\mathcal{G}'$.  □

## Proofs Related to Section 4.2 (Under the Preferred Semantics)

**Proof of Proposition 14:** If $Z$ is not attacked by $\mathcal{G}$, then $Z$ is not attacked in $\mathcal{G}'$. So, due to Proposition 1, each preferred extension of $\mathcal{G}'$ contains $Z$.  □

**Proof of Proposition 15:**

- $\mathcal{E}_i$ is conflict-free in $\mathcal{G}$, so also in $\mathcal{G}'$. Let $A \in \mathcal{E}_i$ being attacked in $\mathcal{G}'$. As $Z$ does not attack $\mathcal{E}_i$, $A$ is attacked in $\mathcal{G}$ and as $\mathcal{E}_i$ is admissible in $\mathcal{G}$, $\mathcal{E}_i$ defends $A$. So, $\mathcal{E}_i$ remains admissible in $\mathcal{G}'$.

- $Z$ does not attack $\mathcal{E}_i$, and $\mathcal{E}_i$ defends $Z$, so $\mathcal{E}_i$ does not attack $Z$ and then $\mathcal{E}_i \cup \{Z\}$ is conflict-free in $\mathcal{G}'$. Let $A \in \mathcal{E}_i \cup \{Z\}$ being attacked in $\mathcal{G}'$. Either $A \in \mathcal{E}_i$ and as we have just proved in the first item that $\mathcal{E}_i$ is admissible in $\mathcal{G}'$, $\mathcal{E}_i$ defends $A$. Or $A = Z$, and we have assumed that $\mathcal{E}_i$ defends $Z$. In each case, $\mathcal{E}_i \cup \{Z\}$ is admissible in $\mathcal{G}'$.

<div align="right">□</div>

**Lemma 2** *If $\mathcal{E}'_i$ is an extension of $\mathcal{G}'$ not containing $Z$, then $\mathcal{E}'_i$ is admissible in $\mathcal{G}$.*

**Proof of Lemma 2:**

As $\mathcal{E}_i'$ does not contain $Z$, $\mathcal{E}_i' \subseteq \mathcal{G}$. $\mathcal{E}_i'$ is conflict-free in $\mathcal{G}'$ so $\mathcal{E}_i'$ is also conflict-free in $\mathcal{G}$. Let $Y \in \mathcal{E}_i'$ being attacked by an argument $A$, $A \in \mathcal{G}$. As $\mathcal{E}_i'$ is admissible, it defends $Y$. So, there is an argument $B \in \mathcal{E}_i'$ attacking $A$. As $\mathcal{E}_i' \subseteq \mathcal{G}$, $B \in \mathcal{G}$. So, we have proved that $\mathcal{E}_i'$ is admissible in $\mathcal{G}$. □

**Proof of Corollary 3:** From Proposition 15, $\forall \mathcal{E}_i \in \mathbf{E}$, $\mathcal{E}_i \cup \{Z\}$ is admissible in $\mathcal{G}'$. So, there exists $j \geq 1$ such that $\mathcal{E}_i \cup \{Z\} \subseteq \mathcal{E}_j'$. From Lemma 2, if $\mathcal{E}_k'$ is an extension of $\mathcal{G}'$ not containing $Z$, then $\mathcal{E}_k'$ is admissible in $\mathcal{G}$. So, there exists $i \geq 1$ such that $\mathcal{E}_k' \subseteq \mathcal{E}_i$. So, we have $\mathcal{E}_k' \subseteq \mathcal{E}_i \subset \mathcal{E}_i \cup \{Z\} \subseteq \mathcal{E}_j'$. As a consequence, there would be a strict inclusion between two extensions of $\mathcal{G}'$, which is impossible. So, there cannot exist $\mathcal{E}_k'$ extension of $\mathcal{G}'$ not containing $Z$, and so each extension of $\mathcal{G}'$ contains $Z$. □

**Proof of Proposition 16:** If $Z$ is not attacked by $\mathcal{G}$, then $Z$ is not attacked in $\mathcal{G}'$ and $Z$ belongs to each preferred extension. Moreover, if there is no even-length cycle in $\mathcal{G}$, due to Lemma 1, there is no even-length cycle in $\mathcal{G}'$. So, due to Proposition 2.5, $\mathcal{G}'$ has only one preferred extension which is not empty (it contains at least $Z$). □

**Lemma 3** *If $Z$ attacks no argument of $\mathcal{G}$ and $\mathcal{E}_i'$ is a non empty extension of $\mathcal{G}'$, then $\mathcal{E}_i' \setminus \{Z\}$ is admissible in $\mathcal{G}$.*

**Proof of Lemma 3:**

- $\mathcal{E}_i'$ is conflict-free in $\mathcal{G}'$ so $\mathcal{E}_i' \setminus \{Z\}$ is also conflict-free in $\mathcal{G}'$ and in $\mathcal{G}$.

- Let $Y \in \mathcal{E}_i' \setminus \{Z\}$. Assume that there is an argument $A$ attacking $Y$. Then $A \neq Z$ since $Z$ attacks no argument of $\mathcal{G}$. $\mathcal{E}_i'$ is a non-empty preferred extension of $\mathcal{G}'$, so there is an argument $B \in \mathcal{E}_i'$ attacking $A$, and $B \neq Z$ (always because $Z$ attacks no argument of $\mathcal{G}$). So, we have $B \in \mathcal{E}_i' \setminus \{Z\}$, and $\mathcal{E}_i' \setminus \{Z\}$ defends $Y$. So, $\mathcal{E}_i' \setminus \{Z\}$ is admissible in $\mathcal{G}$.

□

**Proof of Proposition 17:** Suppose that $Z$ attacks no argument of $\mathcal{G}$ and $\mathbf{E} = \{\{\}\}$.

(*reductio ad absurdum*): Assume that there exists a non-empty extension of $\mathcal{G}'$ denoted by $\mathcal{E}'$. So there exists $Y$ such that $Y \in \mathcal{E}'$. Either $Y = Z$, or $Y \in \mathcal{G}$. In both cases, $Y$ is attacked, because all arguments of $\mathcal{G}$ are attacked (since $\mathbf{E} = \{\{\}\}$) and $Z$ attacks no argument of $\mathcal{G}$. So $\mathcal{E}'$ must defend $Y$. If $Y = Z$, $\mathcal{E}'$ cannot be reduced to $Y$ (because $Z$ attacks no argument and cannot defend itself). So $\mathcal{E}' \setminus \{Z\} \neq \{\}$. If $Y \neq Z$, $Y \in \mathcal{E}' \setminus \{Z\}$, and $\mathcal{E}' \setminus \{Z\} \neq \{\}$. Due to Lemma 3, $\mathcal{E}' \setminus \{Z\}$ is admissible in $\mathcal{G}$ and so $\mathcal{E}' \setminus \{Z\} \subseteq \mathcal{E}$ with $\mathcal{E}$ being a preferred extension of $\mathcal{G}$. So $\mathcal{G}$ has a non-empty extension, which is in contradiction with the assumption. □

**Proof of Proposition 18:**

- $Z$ attacks no argument of $\mathcal{G}$, so due to Proposition 15, $\forall i$, $\mathcal{E}_i$ is admissible in $\mathcal{G}'$. So there exists a preferred extension $\mathcal{E}_j'$ of $\mathcal{G}'$ including $\mathcal{E}_i$. $\mathbf{E} \neq \{\{\}\}$, so $\forall i$, $\mathcal{E}_i \neq \{\}$ and so $\mathcal{E}_j' \neq \{\}$. But $Z \notin \mathcal{E}_i$, so $\mathcal{E}_i \subseteq \mathcal{E}_j' \setminus \{Z\}$. Due to Lemma 3, $\mathcal{E}_j' \setminus \{Z\}$ is admissible in $\mathcal{G}$, so there exists $k \geq 1$ such that $\mathcal{E}_i \subseteq \mathcal{E}_j' \setminus \{Z\} \subseteq \mathcal{E}_k$. Using the definition of a preferred extension ($\subseteq$-maximal among the admissible sets), we can conclude that $\mathcal{E}_i = \mathcal{E}_j' \setminus \{Z\} = \mathcal{E}_k$. So, either $\mathcal{E}_j' = \mathcal{E}_i$ (if $Z \notin \mathcal{E}_j'$), or $\mathcal{E}_j' = \mathcal{E}_i \cup \{Z\}$ (if $Z \in \mathcal{E}_j'$). In the first case, $\mathcal{E}_i$ is an extension of $\mathcal{G}'$. In the second case, $\mathcal{E}_i \cup \{Z\}$ is an extension of $\mathcal{G}'$. Moreover, if $Z \in \mathcal{E}_j'$, $\mathcal{E}_j'$ defends $Z$ (which is attacked by $\mathcal{G}$, since it does not attack $\mathcal{G}$) and as $Z$ attacks no argument, $\mathcal{E}_i = \mathcal{E}_j' \setminus \{Z\}$ defends $Z$. So, if $\mathcal{E}_i$ does not defend $Z$, $\mathcal{E}_i$ is an extension of $\mathcal{G}'$. On the other hand, if $\mathcal{E}_i$ defends $Z$, $\mathcal{E}_i \cup \{Z\}$ is conflict-free in $\mathcal{G}'$. So, $\mathcal{E}_i \cup \{Z\}$ is admissible in $\mathcal{G}'$ and it is the case when $Z \in \mathcal{E}_j'$ and $\mathcal{E}_i \cup \{Z\}$ is an extension of $\mathcal{G}'$.

- Now, we prove that $\mathcal{G}$ and $\mathcal{G}'$ have the same number of extensions. From the first part of the proof, we know that each extension of $\mathcal{G}$ is included in an extension of $\mathcal{G}'$. Moreover, two distinct extensions of $\mathcal{G}$ cannot be included in a same extension of $\mathcal{G}'$. Indeed, the union of two non-empty preferred extensions defends all its elements and strictly contains each of these extensions. So the union of two extensions cannot be conflict-free.

So, we know that $\mathcal{G}'$ has at least as many extensions as $\mathcal{G}$ and that $\mathcal{G}'$ has at least one non-empty extension. So, $\forall \mathcal{E}_j' \in \mathbf{E}'$, $\mathcal{E}_j' \neq \{\}$. Due to Lemma 3, $\mathcal{E}_j' \setminus \{Z\}$ is admissible in $\mathcal{G}$. So, for each $\mathcal{E}_j'$, there exists $\mathcal{E}_i$, an extension of $\mathcal{G}$ such that $\mathcal{E}_j' \setminus \{Z\} \subseteq \mathcal{E}_i$. From the first part of the proof, we have:

  - Either $\mathcal{E}_i$ defends $Z$, and $\mathcal{E}_i \cup \{Z\}$ is an extension of $\mathcal{G}'$. As $\mathcal{E}_j' \setminus \{Z\} \subseteq \mathcal{E}_i$, we have $\mathcal{E}_j' \subseteq \mathcal{E}_i \cup \{Z\}$, and as $\mathcal{E}_j'$ is maximal admissible in $\mathcal{G}'$, $\mathcal{E}_j' = \mathcal{E}_i \cup \{Z\}$.
  - Or $\mathcal{E}_i$ does not defend $Z$, $\mathcal{E}_i$ is an extension of $\mathcal{G}'$. As $\mathcal{E}_j'$ is maximal admissible in $\mathcal{G}'$, we have $Z \notin \mathcal{E}_j'$ and $\mathcal{E}_j' = \mathcal{E}_j' \setminus \{Z\} = \mathcal{E}_i$.

So, $\mathcal{G}$ and $\mathcal{G}'$ have the same number of extensions.

□

**Proof of Proposition 19:**

- If $\mathbf{E} = \{\{\}\}$, obviously each change satisfies Monotony.
- If $\mathcal{G}$ has a non-empty extension, Proposition 15 can be applied. So each extension of $\mathcal{G}$ remains admissible in $\mathcal{G}'$ and is included in a preferred extension of $\mathcal{G}'$. So, the change satisfies Monotony.

□

**Proof of Proposition 20:** Let $\mathcal{E} = \cap_{i \geq 1} \mathcal{E}_i$ and $\mathcal{E}' = \cap_{i \geq 1} \mathcal{E}_i'$.

Let $\mathcal{E}_g$ (resp. $\mathcal{E}_g'$) denote the grounded extension of $\mathcal{G}$ (resp $\mathcal{G}'$). Due to Proposition 1.4, we know that $\mathcal{E}_g \subseteq \mathcal{E}$ and $\mathcal{E}_g' \subseteq \mathcal{E}'$. Dung (1995) has proved that when there is no controversial argument, the grounded extension is exactly the intersection of the preferred extensions. So, if $\mathcal{G}$ contains no controversial argument, we have $\mathcal{E}_g = \mathcal{E}$.

Now, if $Z$ does not attack $\cap_{i \geq 1} \mathcal{E}_i$, $Z$ does not attack $\mathcal{E}_g$, so due to Proposition 10, if $\mathcal{E} \neq \{\}$ then we have $\mathcal{E}_g \subseteq \mathcal{E}_g'$ when $\mathcal{E} = \{\}$ then $\cap_{i \geq 1} \mathcal{E}_i = \{\}$ and the inclusion trivially holds. So, we have $\mathcal{E} = \mathcal{E}_g \subseteq \mathcal{E}_g' \subseteq \mathcal{E}'$, and then $\cap_{i \geq 1} \mathcal{E}_i \subseteq \cap_{i \geq 1} \mathcal{E}_i'$. □

**Proof of Proposition 21:** $\mathbf{E} \neq \{\{\}\}$ and there is no even-length cycle in $\mathcal{G}'$ so there is no even-length cycle in $\mathcal{G}$; as a consequence, according to Proposition 2.5 there is only one extension $\mathcal{E}$ in $\mathcal{G}$; moreover, $\mathcal{E} \neq \{\}$. Since there is no even-length cycle in $\mathcal{G}'$, we know that there is only one extension $\mathcal{E}'$ in $\mathcal{G}'$. Assume that $Z$ and each unattacked argument $A_i$ in $\mathcal{G}$ are attacked in $\mathcal{G}'$; so there is no unattacked argument in $\mathcal{G}'$.

Assume that $\mathcal{E}' \neq \{\}$. Let $X \in \mathcal{E}'$. $X$ is attacked in $\mathcal{G}'$. Let $Y_1$ denote an attacker of $X$. As $\mathcal{E}'$ is admissible, $\mathcal{E}'$ defends $X$. So $\mathcal{E}'$ contains $X_2$ which attacks $Y_1$. As there is no even-length cycle in $\mathcal{G}'$, we know that $X_2 \neq X$. And $X_2$ is not unattacked.

So we are able to built an infinite sequence of distinct arguments:

$X$ is attacked by $Y_1$ attacked by $X_2 \ldots Y_p$ attacked by $X_{p+1}$ attacked by $Y_{p+1} \ldots$

The $X_i$'s (resp. $Y_i$'s) are distinct due to the absence of even-length cycles in $\mathcal{G}'$.

It contradicts the assumption that $\mathbf{A}$ is finite. So $\mathcal{E}' = \{\}$ and the change is destructive. □

## Appendix B. Illustration of Properties for the Other Change Operations

The following examples illustrate the structural properties and the property of Monotony for change operations distinct from $\oplus_i^a$ (let us recall that the property of Priority to Recency does not make sense for these other change operations).

**For $\ominus_i^a$**

First, we notice that if $\langle \mathbf{A}, \mathbf{R} \rangle \oplus_i^a (Z, \mathcal{I}_z) = \langle \mathbf{A}', \mathbf{R}' \rangle$ then $\langle \mathbf{A}', \mathbf{R}' \rangle \ominus_i^a Z = \langle \mathbf{A}, \mathbf{R} \rangle$. So in each example of Section 3.2, a change $\ominus_i^a$ can also be illustrated.

- Example 4.1 show a decisive change $\ominus_i^a$ and a change $\ominus_i^a$ which does not satisfy the property of Monotony.

  Example 5.2 shows a decisive change $\ominus_i^a$ and a change $\ominus_i^a$ which satisfies the property of Monotony .

- Example 4.3 shows a restrictive change $\ominus_i^a$ and a change $\ominus_i^a$ which does not satisfy the property of Monotony

- Examples 2.3, 3.1, 5.1 show a questioning change $\ominus_i^a$ and a change $\ominus_i^a$ which does not satisfy the property of Monotony

- Examples 2.1, 2.2, 4.2 show a destructive change $\ominus_i^a$ and a change $\ominus_i^a$ which does not satisfy the property of Monotony

- Example 8.2 shows an expansive change $\ominus_i^a$ and a change $\ominus_i^a$ which satisfies the property of Monotony

- Examples 7.1, 7.2, 7.3 show a conservative change $\ominus_i^a$ and a change $\ominus_i^a$ which satisfies the property of Monotony

- Examples 6, 8.1 show an altering change $\ominus_i^a$ and a change $\ominus_i^a$ which does not satisfy the property of Monotony

**For $\oplus_i$ and $\ominus_i$**

- With $\langle \mathbf{A} = \{A, B, C\}, \mathbf{R} = \{(A, B), (B, C), (C, A)\} \rangle$, $\langle \mathbf{A}, \mathbf{R} \rangle \oplus_i (A, C)$ is a decisive change (before the change $\mathbf{E} = \{\{\}\}$, after the change $\mathbf{E}' = \{\{A\}\}$); and the inverse operation $\langle \mathbf{A}, \mathbf{R} \cup \{(A, C)\} \rangle \ominus_i (A, C)$ is destructive.

  In this example, $\oplus_i$ satisfies the property of Monotony and $\ominus_i$ does not.

- With $\langle \mathbf{A} = \{A, B, C\}, \mathbf{R} = \{(A, B), (B, C)\} \rangle$, $\langle \mathbf{A}, \mathbf{R} \rangle \oplus_i (C, A)$ is a destructive change (before the change $\mathbf{E} = \{\{A, C\}\}$, after the change $\mathbf{E}' = \{\{\}\}$); and the inverse operation $\langle \mathbf{A}, \mathbf{R} \cup \{(C, A)\} \rangle \ominus_i (C, A)$ is decisive.

  In this example, $\oplus_i$ does not satisfy the property of Monotony and $\ominus_i$ satisfies it.

- With $\langle \mathbf{A} = \{A, B, C\}, \mathbf{R} = \{(A, B), (B, C)\} \rangle$, $\langle \mathbf{A}, \mathbf{R} \rangle \oplus_i (A, C)$ is an altering change (before the change $\mathbf{E} = \{\{A, C\}\}$, after the change $\mathbf{E}' = \{\{A\}\}$); and the inverse operation $\langle \mathbf{A}, \mathbf{R} \cup \{(A, C)\} \rangle \ominus_i (A, C)$ is expansive.

In this example, $\oplus_i$ does not satisfy the property of Monotony and $\ominus_i$ satisfies it.

- With $\langle \mathbf{A} = \{A, B, C\}, \mathbf{R} = \{(A, B)\}\rangle$, $\langle \mathbf{A}, \mathbf{R}\rangle \oplus_i (C, B)$ is a conservative change (before the change $\mathbf{E} = \{\{A, C\}\}$, after the change $\mathbf{E}' = \{\{A, C\}\}$); and the inverse operation $\langle \mathbf{A}, \mathbf{R} \cup \{(A, C)\}\rangle \ominus_i (A, C)$ is conservative.

  In this example, $\oplus_i$ and $\ominus_i$ satisfy the property of Monotony.

- With $\langle \mathbf{A} = \{A, B, C, D\}, \mathbf{R} = \{(A, B), (B, A), (B, C), (D, C)\}\rangle$, $\langle \mathbf{A}, \mathbf{R}\rangle \oplus_i (C, D)$ is a questioning change (before the change $\mathbf{E} = \{\{A, D\}, \{B, D\}\}$, after the change $\mathbf{E}' = \{\{A, D\}, \{B, D\}, \{A, C\}\}$); and the inverse operation $\langle \mathbf{A}, \mathbf{R} \cup \{(C, D)\}\rangle \ominus_i (C, D)$ is restrictive.

  In this example, $\oplus_i$ satisfies the property of Monotony and $\ominus_i$ does not.

- With $\langle \mathbf{A} = \{A, B, C, D\}, \mathbf{R} = \{(A, B), (B, A), (B, C), (D, C), (C, D)\}\rangle$, the change $\langle \mathbf{A}, \mathbf{R}\rangle \oplus_i (A, D)$ is a restrictive one (before the change $\mathbf{E} = \{\{A, D\}, \{B, D\}, \{A, C\}\}$, after the change $\mathbf{E}' = \{\{B, D\}, \{A, C\}\}$).

  The inverse operation $\langle \mathbf{A}, \mathbf{R} \cup \{(A, D)\}\rangle \ominus_i (A, D)$ is questioning.

  In this example, $\oplus_i$ does not satisfy the property of Monotony and $\ominus_i$ satisfies it.

- With $\langle \mathbf{A} = \{A, B, C\}, \mathbf{R} = \{(A, B), (B, C)\}\rangle$, $\langle \mathbf{A}, \mathbf{R}\rangle \ominus_i (A, B)$ is an altering change (before the change $\mathbf{E} = \{\{A, C\}\}$, after the change $\mathbf{E}' = \{\{A, B\}\}$).

  In this example, $\ominus_i$ does not satisfy the property of Monotony.

- With $\langle \mathbf{A} = \{A, B, C, D\}, \mathbf{R} = \{(A, B), (B, C), (C, A)\}\rangle$, $\langle \mathbf{A}, \mathbf{R}\rangle \oplus_i (D, A)$ is a expansive change (before the change $\mathbf{E} = \{\{D\}\}$, after the change $\mathbf{E}' = \{\{D, B\}\}$).

  In this example, $\oplus_i$ satisfies the property of Monotony.